\documentclass[10pt,twocolumn,letterpaper]{article}

\PassOptionsToPackage{table}{xcolor}

\usepackage{wacv} %

\usepackage{multirow}
\usepackage{pifont}

\newsavebox{\fitbox}
\newcommand{\fitwidth}[2][\textwidth]{%
    \sbox{\fitbox}{#2}%
    \ifdim\wd\fitbox>#1
        \resizebox{#1}{!}{\usebox{\fitbox}}%
    \else
        \usebox{\fitbox}%
    \fi
}

\usepackage{needspace} %
\usepackage[accsupp]{axessibility}
\definecolor{wacvblue}{rgb}{0.21,0.49,0.74}
\usepackage[pagebackref,breaklinks,colorlinks,allcolors=wacvblue]{hyperref}

\def\wacvPaperID{211} %
\def\confName{WACV}
\def\confYear{2027}

\title{TMAL: Topology-preserving Multi-offset Affinity Learning \\for Curvilinear Structure Segmentation}

\author{Suho Han$^{1*}$, Soungsill Park$^{1*}$, Kanghyun Ryu$^{2\dagger}$, and Yoseob Han$^{1\dagger}$\\
$^{1}$Soongsil University, Republic of Korea\\
$^{2}$Korea Institute of Science and Technology, Republic of Korea\\
{\tt\small suhohan@soongsil.ac.kr, sspark@ssu.ac.kr, khryu@kist.re.kr, yoseob.han@ssu.ac.kr}\\
{\small $^{*}$Equal contribution \quad $^{\dagger}$Co-corresponding authors}
}

\begin{document}
\maketitle
\begin{abstract}
    Curvilinear structures such as retinal vessels and surface cracks need both pixel-level accuracy and structural continuity.
    Pixel-wise supervision does not model the connections between pixels, and topology-based losses still produce ordinary masks.
    Connectivity-based methods predict these connections directly, but they use binary connectivity targets over a fixed local range.
    Such labels give the same value to thin segments and to thick regions, and they cannot describe connections at larger offsets.
    To address these limitations, we propose \textbf{Topology-preserving Multi-offset Affinity Learning (TMAL)}.
    TMAL uses soft affinities that emphasize thin regions near the boundary, and it predicts directional affinities at several offsets to reach further.
    It then combines these affinities with weights that come from the decoder features, so local continuity and wider context are used together.
    An auxiliary segmentation branch adds region-level supervision and keeps the overlap accuracy high.
    On the retinal vessel dataset DRIVE and the crack inspection dataset DeepCrack,
    TMAL reaches the highest Dice, IoU, and clDice scores, and it greatly reduces the connected-component error $\Delta\beta_0$.
    Code is available at \url{https://github.com/suho-han/TMAL}.
\end{abstract}

\section{Introduction}
\label{sec:intro}

Curvilinear structure segmentation finds thin and long patterns, such as blood vessels in medical images and cracks in industrial images.
Good segmentation needs both pixel-level accuracy and structural continuity.
Even a small local error can break a thin branch and split one structure into several pieces.
Such errors then harm later tasks that assume continuous paths, such as vessel analysis and crack inspection.

Most methods treat segmentation as pixel-wise classification.
They label each pixel as foreground or background and optimize binary cross-entropy and Dice losses~\cite{milletari2016v}.
These losses improve local accuracy and region overlap, but they do not supervise the relations between foreground pixels.
Therefore, a prediction can have high pixel accuracy and still contain broken branches.
Topology-based methods attack this problem through the training objective.
They reward centerline preservation~\cite{shit2021cldice} or low-cost paths between components~\cite{esmaeilzadeh2025cape}.
However, their outputs are still ordinary masks and do not represent pixel-to-pixel connections.

Connectivity-based methods predict directional connections between pixels~\cite{yang2022biconnet,yang2023directional}.
Their targets are binary, so every positive connection receives the same value.
This treats thin segments like thick ones, even though thin segments break more easily.
These methods also read a fixed local neighborhood, usually at one-pixel offsets, so they cannot describe connections at larger offsets.

\begin{figure*}[!t]
    \centering
    \includegraphics[width=\textwidth]{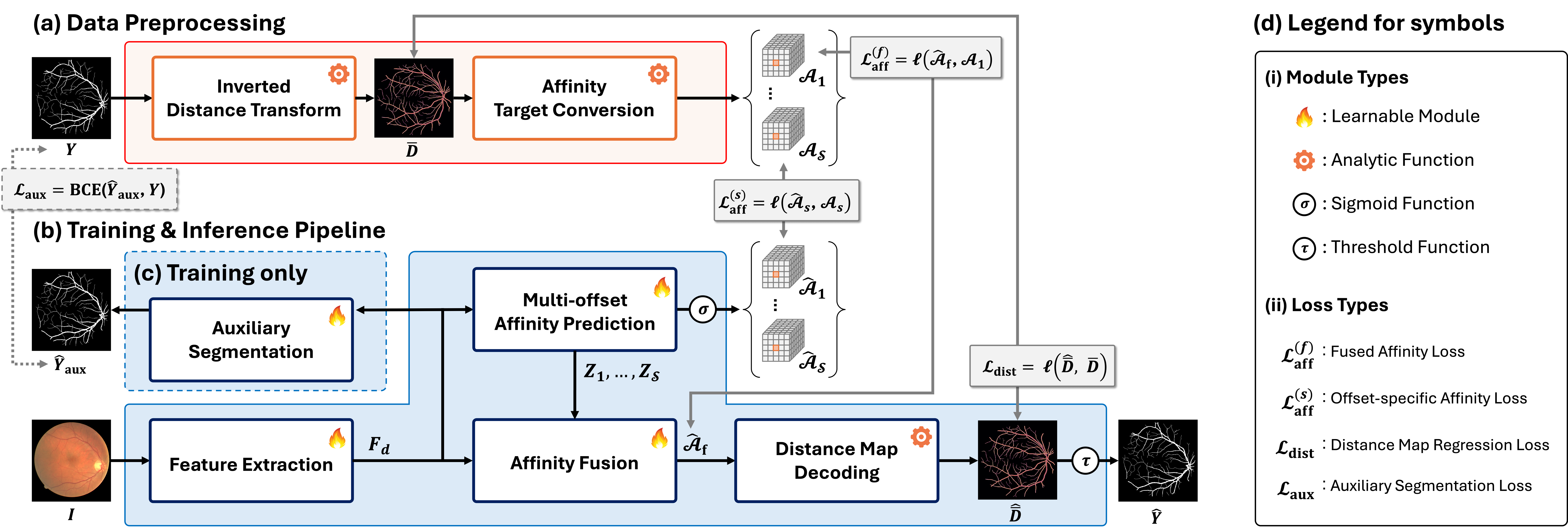}
    \caption[Overview of TMAL]{
    Overview of the proposed TMAL framework.
    (a) The Inverted Distance Transform (IDT) maps the binary ground-truth mask $Y$ to the inverted distance map $\bar{\mathcal{D}}$, from which Affinity Target Conversion (ATC) derives the targets $\{\mathcal{A}_{s}\}_{s\in\mathcal{S}}$.
    (b) Feature Extraction (FE) maps the input image $I$ to the decoder feature $F_\mathrm{d}$, which Multi-offset Affinity Prediction (MAP) and Affinity Fusion (AF) turn into the fused affinity $\hat{\mathcal{A}}_{\mathrm{f}}$, and Distance Map Decoding (DMD) into $\hat{\bar{\mathcal{D}}}$ and the mask $\hat{Y}$.
    (c) Auxiliary Segmentation (AuxSeg), used only during training, predicts $\hat{Y}_{\mathrm{aux}}$ from $F_\mathrm{d}$.
    (d) Legend for the module and loss symbols.
    }
    \label{fig:overview}
\end{figure*}

To overcome these limitations, we propose \textbf{Topology-preserving Multi-offset Affinity Learning (TMAL)}, shown in Figure~\ref{fig:overview}.
It replaces binary connectivity targets with soft affinities from the \textbf{Inverted Distance Transform (IDT)}.
This target measures how close a pixel is to the boundary and how thick the structure is, so it gives larger values to thin regions near the boundary, where breaks are more likely.
\textbf{Multi-offset Affinity Prediction (MAP)} predicts directional affinities at several offsets, and \textbf{Affinity Fusion (AF)} weights them adaptively from the decoder features.
\textbf{Auxiliary Segmentation (AuxSeg)} adds pixel-level supervision during training, so the model does not trade region-level accuracy for connectivity.

Our main contributions are summarized as follows:
\begin{itemize}
    \item We replace binary connectivity targets with soft directional affinities derived from boundary distance, with larger values in thin, boundary-adjacent regions.
    \item We design MAP to predict affinities at several sampling offsets and AF to combine them with adaptive weights.
    \item TMAL improves overlap and skeleton agreement, and it substantially reduces the connected-component error $\Delta\beta_0$ on a medical dataset, DRIVE~\cite{staal2004ridge} and an industrial dataset, DeepCrack~\cite{liu2019deepcrack}.
\end{itemize}

\section{Related Work}

\paragraph{Mask-based Approaches.}
Encoder-decoder networks are still the main design for dense prediction.
\textbf{U-Net}~\cite{ronneberger2015u} set the pattern with a contracting path for context and a symmetric expanding path for localization.
Later variants change how the decoder selects and combines features, such as the attention gates of \textbf{Attention U-Net}~\cite{oktay2018attention} and the channel and spatial attention of \textbf{CS-Net}~\cite{mou2019cs}.
Others change the encoder or the output stage, such as the context encoder of \textbf{CE-Net}~\cite{gu2019cenet} and the full-resolution aggregation and dual-threshold recovery of \textbf{FR-UNet}~\cite{liu2022full}.
Generative methods instead model the mask as a distribution.
GANs regularize that distribution~\cite{son2019towards}, and \textbf{SegDiff}~\cite{amit2021segdiff} and \textbf{MedSegDiff}~\cite{wu2024medsegdiff} denoise it with an image-conditioned DDPM~\cite{ho2020denoising}.
All of these methods produce pixel-wise masks and never represent the relations between neighboring pixels.

\paragraph{Distance-based Approaches.}
Distance-based methods build their targets from distance transforms.
\textbf{Shape-aware Segmentation}~\cite{xue2020shape} predicts signed distance maps as a global shape prior.
\textbf{Boundary Loss}~\cite{kervadec2021boundary} instead weights its objective by the distance to the ground-truth contour.
More recently, the \textbf{SAUNA transform}~\cite{dang2025imagelevel} encodes annotation uncertainty from boundary proximity and vessel thickness for retinal vessels.
All of these methods attach distance information to single pixels.
We instead use it to describe directional relations between pairs of pixels, and we put more weight on connectivity in thin structures.

\paragraph{Topology-based Approaches.}
Topology-based methods add a topological term to the training objective.
\textbf{clDice}~\cite{shit2021cldice} measures how well a prediction agrees with its morphological skeleton and gives a differentiable soft form of that measure.
\textbf{CAPE}~\cite{esmaeilzadeh2025cape} picks node pairs from the ground-truth graph and forces low-cost paths between them.
These losses improve topology through supervision alone and never model connectivity between foreground pixels.

\paragraph{Connectivity-based Approaches.}
Recent methods predict connectivity directly in order to model directional relations.
\textbf{BiconNet}~\cite{yang2022biconnet} combines each direction with its opposite by Bilateral Voting.
\textbf{DconnNet}~\cite{yang2023directional} separates directional information across latent channels in its decoder.
Both replace pixel-wise mask prediction with directional connectivity prediction, but their targets are binary and cover only a fixed eight-neighborhood.
Neither method varies the target with local thickness, and that is where thin structures break.

\begin{figure}[!t]
    \centering
    \begin{subfigure}[t]{0.23\textwidth}
        \centering
        \includegraphics[width=\linewidth]{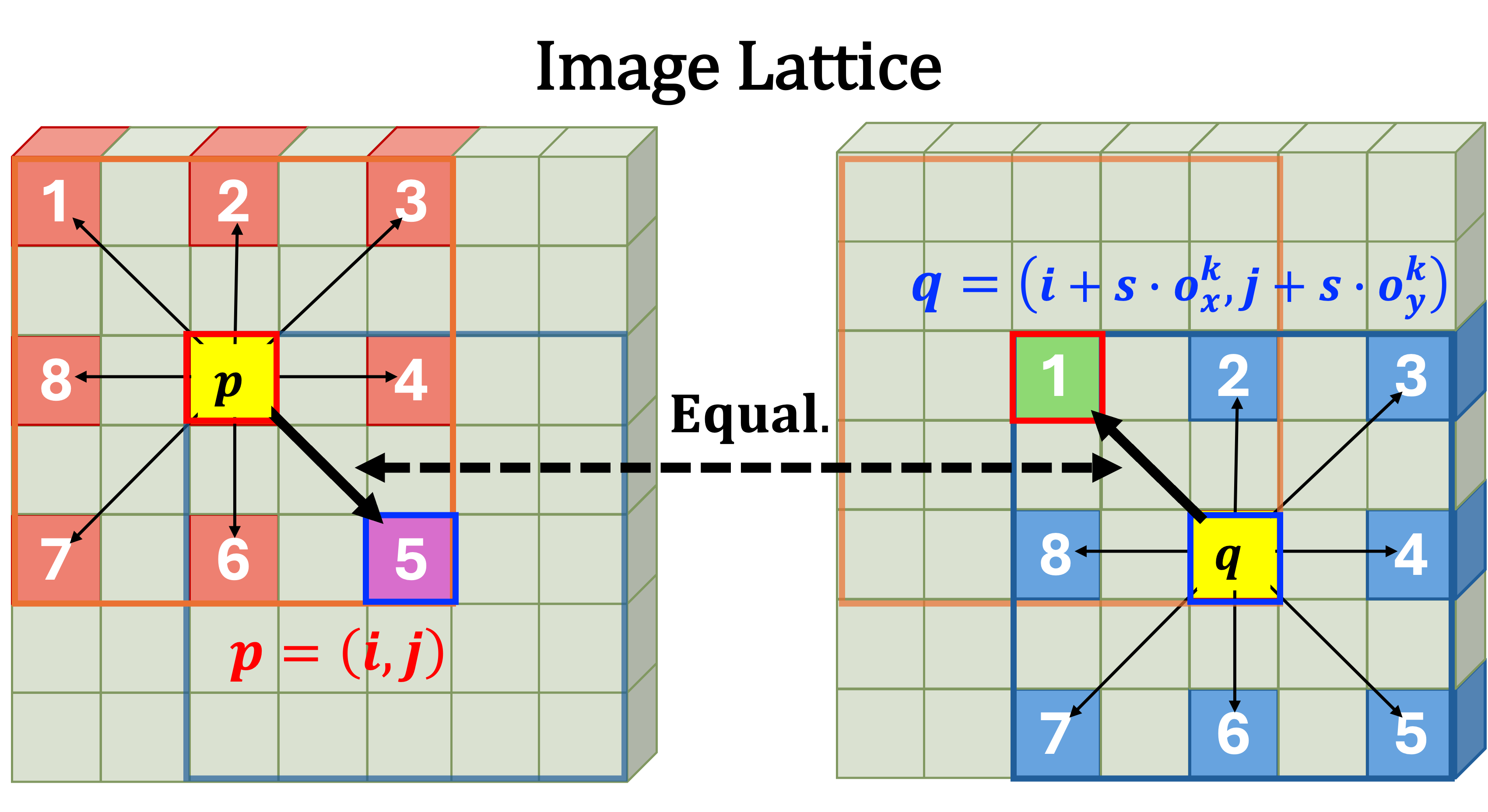}
        \caption{\textbf{Image-domain Perspective}}
    \end{subfigure}
    \hfill
    \begin{subfigure}[t]{0.23\textwidth}
        \centering
        \includegraphics[width=\linewidth]{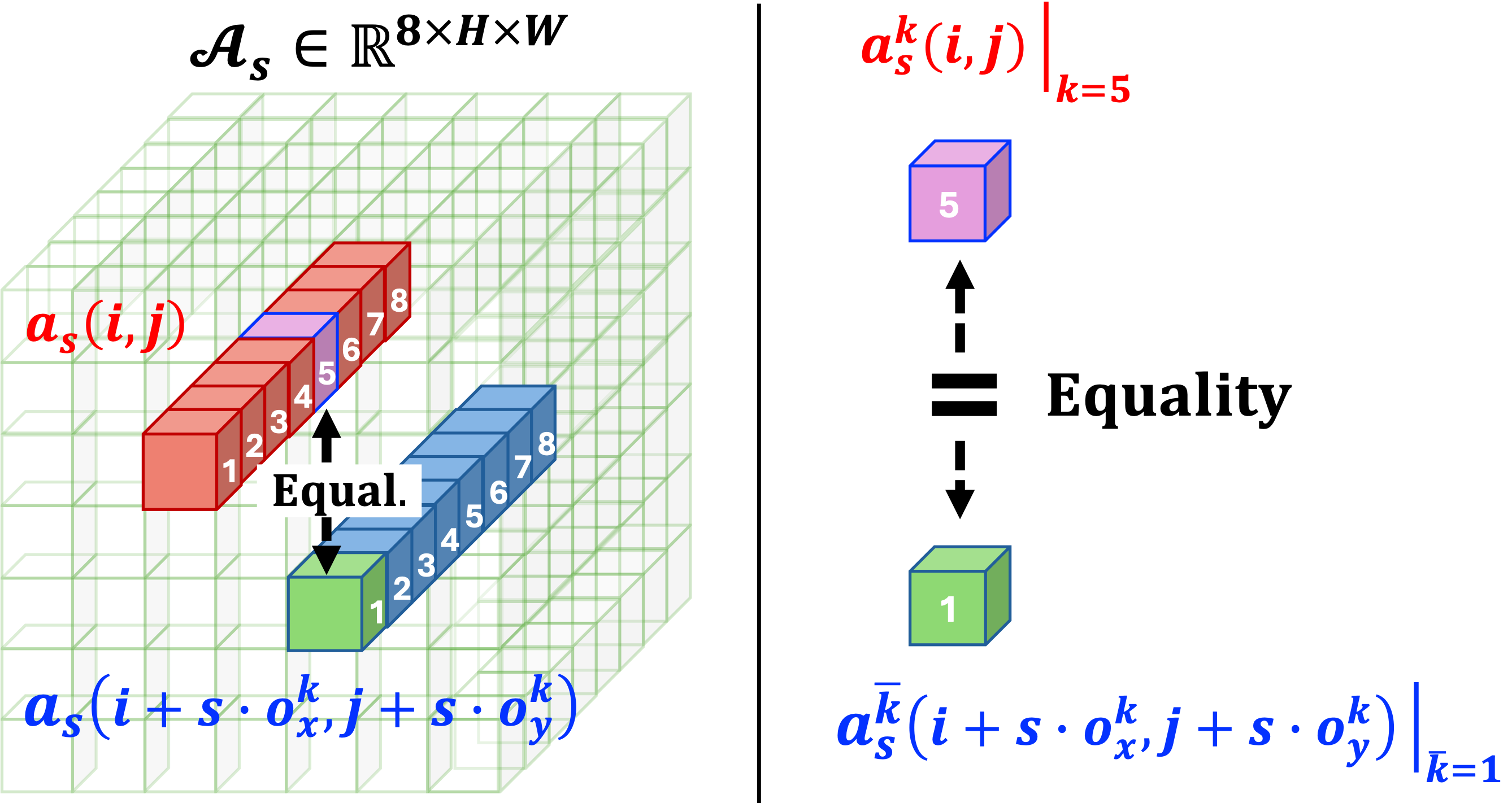}
        \caption{\textbf{Affinity-domain Perspective}}
    \end{subfigure}
    \caption[Directional Pairs]{
    Direction indexing and the symmetry of a directional pair.
    Direction $k$ from $p$ and direction $\bar{k}$ from $q$ select the same pair, seen (a) in the image domain and (b) in the affinity domain.
    (b) writes $a_{s}(i,j)=\{a^{k}_{s}(i,j)\}_{k=1}^{8}$ for the eight-channel affinity vector at pixel $(i,j)$.
    }
    \label{fig:equality}
\end{figure}

\section{Method}
Let $I \in \mathbb{R}^{3 \times H \times W}$ denote the input image, $Y \in \{0,1\}^{H \times W}$ the binary ground-truth mask, and $\hat{Y} \in \{0,1\}^{H \times W}$ the predicted mask.

\subsection{Directional Connectivity Modeling}\label{sec:dcm}
Let $\mathcal{O}=\{(o_x,o_y)\mid o_x,o_y\in\{-1,0,+1\},(o_x,o_y)\neq(0,0)\}$ denote the eight unit directions, taken in the fixed order $\mathcal{O}=\{(o_x^k,o_y^k)\}_{k=1}^{8}$ in Figure~\ref{fig:equality}(a).
We write $\bar{k}$ for the direction opposite to $k$, so that $(o_x^{\bar{k}},o_y^{\bar{k}})=-(o_x^{k},o_y^{k})$.
Each branch carries a direction set of eight elements that is closed under negation.
Branches $s=1$ and $s=2$ use $\mathcal{O}$ itself, and Section~\ref{sec:atc} introduces a rotated set.
A directional pair is given by a direction $k$ and a sampling offset $s$.
The pair links pixel $p=(i,j)$ to pixel $q=(i+s\cdot o_x^k,\,j+s\cdot o_y^k)$.
Here $s$ is a spatial distance, not a change of feature-map resolution.
The pair is unordered, because direction $\bar{k}$ at the same offset links $q$ back to $p$, as shown in Figure~\ref{fig:equality}(a).

Let $y(i,j)$ be the value of $Y$ at pixel $(i,j)$.
Prior connectivity-based models~\cite{yang2022biconnet,yang2023directional} read this pair at the single offset $s=1$, so $q$ is the immediate neighbor of $p$.
They use the binary connectivity target:
\begin{equation}\label{eq:dconn_target}
    c^k(i,j)
    =
    y(i,j)
    \cdot
    y(i+o_x^k,\,j+o_y^k)
    ,
    \quad k=1,\ldots,8.
\end{equation}
This gives $\mathcal{C}=\{c^k\}_{k=1}^{8}\in\{0,1\}^{8\times H\times W}$.
A value of one means that pixel $(i,j)$ and its neighbor in direction $k$ are both foreground.
The target depends only on the two endpoints, so $c^{k}(p)=y(p)\,y(q)=c^{\bar{k}}(q)$ and the two directions of a pair carry the same value, as shown in Figure~\ref{fig:equality}(b).

Conventional connectivity-based models predict $\hat{\mathcal{C}}$
and supervise it against $\mathcal{C}$ with BCE-based objectives.
TMAL keeps this directional pair formulation of Equation~\eqref{eq:dconn_target}, but it replaces the binary value $\mathcal{C}$ with a continuous affinity $\mathcal{A}$ and the single offset $s=1$ with a set of offsets $\mathcal{S}$.

\begin{figure}[!t]
    \centering
    \begin{subfigure}[t]{0.15\textwidth}
        \centering
        \includegraphics[width=\linewidth]{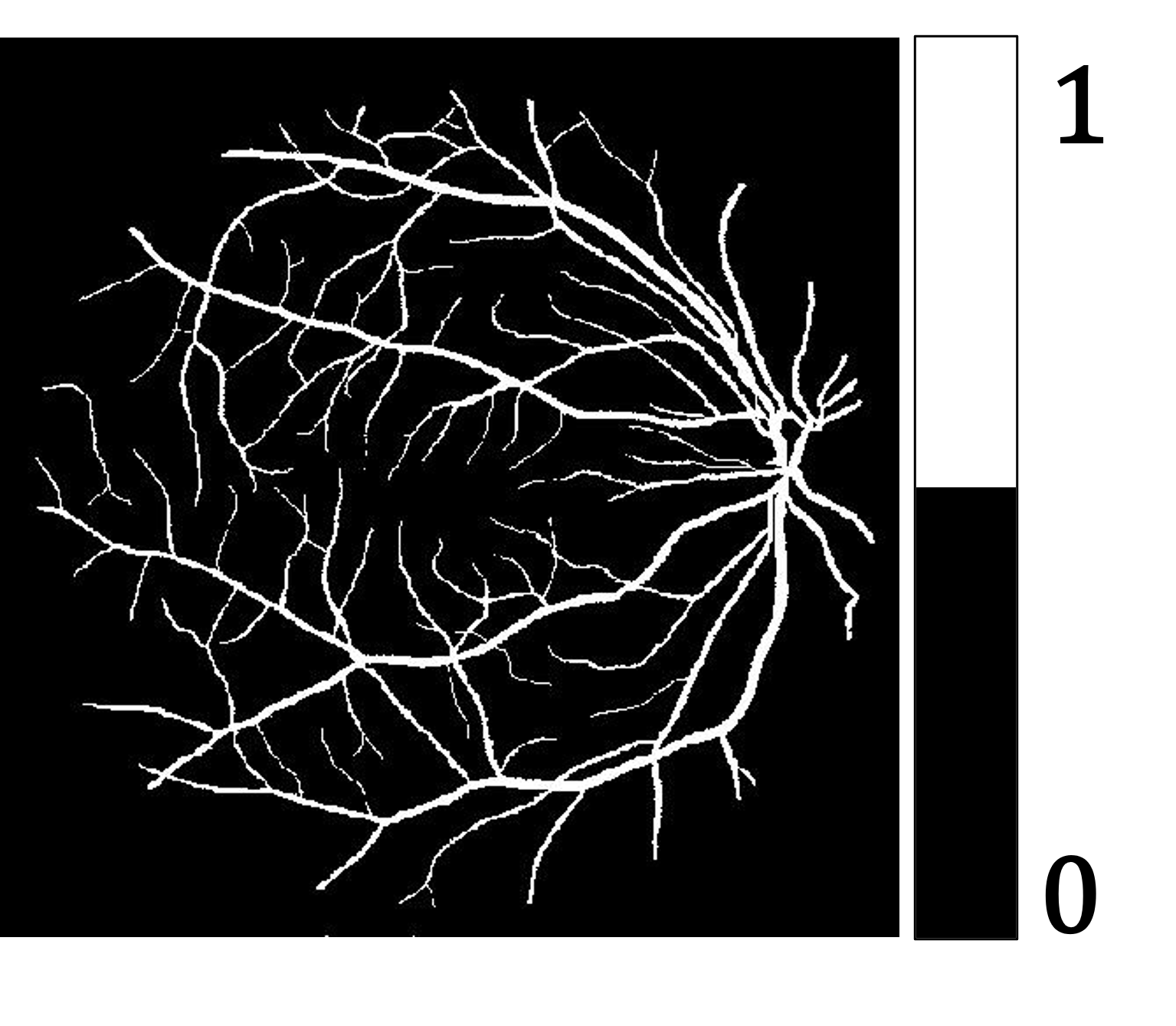}
        \caption{$Y$}
    \end{subfigure}
    \hfill
    \begin{subfigure}[t]{0.15\textwidth}
        \centering
        \includegraphics[width=\linewidth]{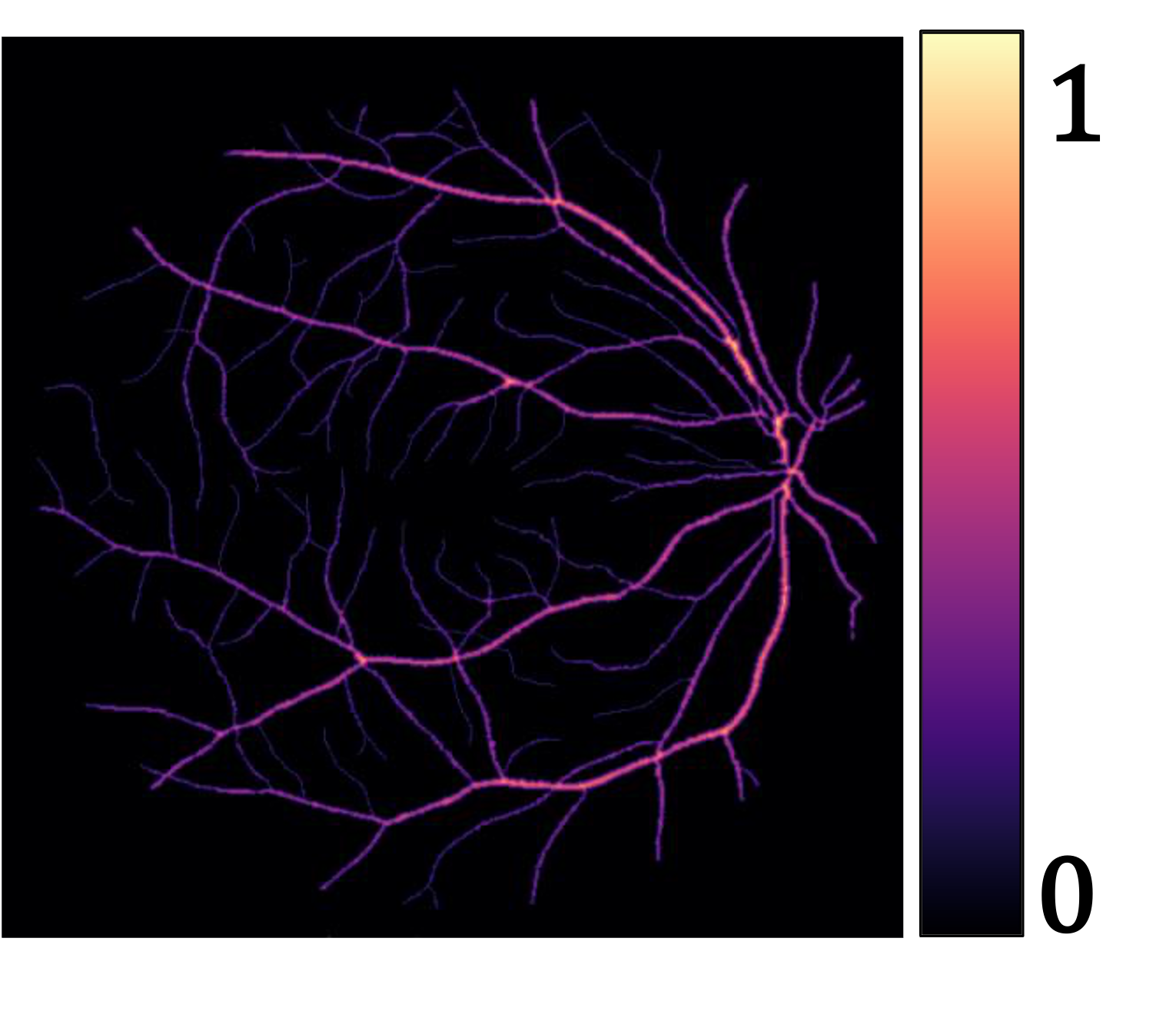}
        \caption{$\mathbf{\mathcal{D}}$}
    \end{subfigure}
    \hfill
    \begin{subfigure}[t]{0.15\textwidth}
        \centering
        \includegraphics[width=\linewidth]{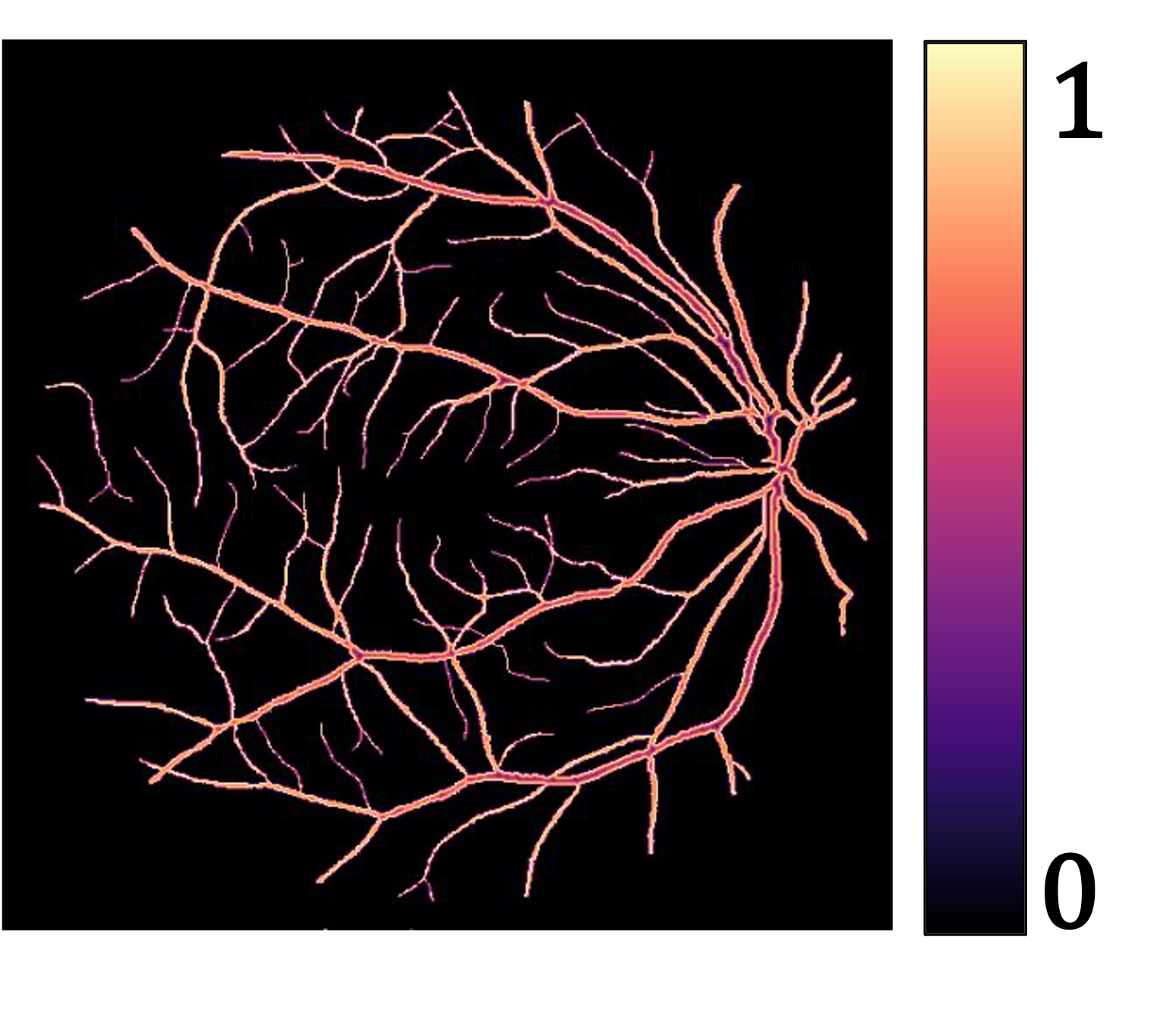}
        \caption{$\bar{\mathbf{\mathcal{D}}}$}
    \end{subfigure}
    \caption[Visualization of the segmentation label]{
        (a) The binary ground-truth mask $Y$, (b) the Euclidean distance map $\mathcal{D}$,
        and (c) the inverted distance map $\bar{\mathcal{D}}$, which peaks on thin branches and near boundaries where $\mathcal{D}$ is smallest.
    } \label{fig:distance_map}
\end{figure}

\subsection{Data Preprocessing}\label{sec:atc}
In Figure~\ref{fig:overview}(a), the IDT converts the binary ground-truth mask $Y$ into a continuous inverted distance map $\bar{\mathcal{D}}$,
and Affinity Target Conversion (ATC) derives multi-offset affinity targets $\{\mathcal{A}_{s}\}_{s\in\mathcal{S}}$ from it.

\paragraph{Inverted Distance Transform.}
Let $\Omega_1=\{(i,j)\mid y(i,j)=1\}$ and $\Omega_0=\{(i,j)\mid y(i,j)=0\}$ denote the foreground and background pixel sets.
The Euclidean Distance Transform first gives each foreground pixel its distance to the nearest background pixel, and zero elsewhere.
\begin{equation}\label{eq:edt}
    \begin{aligned}
         & {}\mathcal{D}(i,j)
        \\
         & = \begin{cases}
                 \displaystyle\min_{(u,v)\in\Omega_0}
                 \sqrt{(i-u)^2+(j-v)^2}, & (i,j)\in\Omega_1, \\
                 0,                      & (i,j)\in\Omega_0.
             \end{cases}
    \end{aligned}
\end{equation}

Then the distance map $\mathcal{D}$ is inverted and normalized so that boundary-adjacent and thin regions receive larger values.
The normalized inverted distance map $\bar{\mathcal{D}}$ is defined as
\begin{equation}\label{eq:inv_norm}
    \bar{\mathcal{D}}(i,j)
    =
    \begin{cases}
        \dfrac{d_{\max}-\mathcal{D}(i,j)}{d_{\max}-d_{\min}}, & (i,j)\in\Omega_1, \\
        0,                                                    & (i,j)\in\Omega_0,
    \end{cases}
\end{equation}
where $d_{\max}=\max_{(i,j)\in\Omega_1}\mathcal{D}(i,j)$ and $d_{\min}=\min_{(i,j)\in\Omega_1}\mathcal{D}(i,j)$.
The resulting map $\bar{\mathcal{D}}\in{[0,1]}^{H\times W}$ is largest in thin regions and near boundaries.
It is smallest inside thick centers, which is the opposite of $\mathcal{D}$, as shown in Figure~\ref{fig:distance_map}.

\begin{figure}[!t]
    \centering
    \begin{subfigure}[t]{0.98\linewidth}
        \centering
        \includegraphics[width=\linewidth]{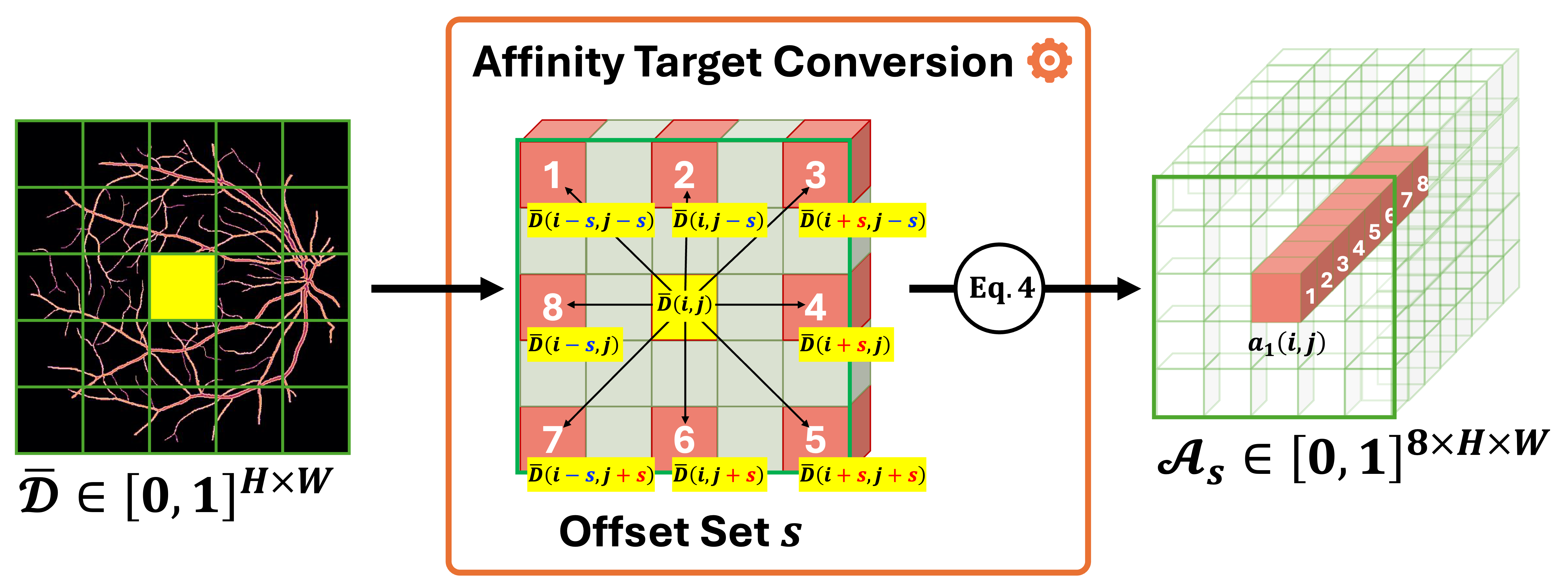}
        \caption{\textbf{Affinity Target Conversion}}
    \end{subfigure}
    \par\vspace{0.5em}
    \begin{subfigure}[t]{0.28\linewidth}
        \centering
        \includegraphics[width=\linewidth]{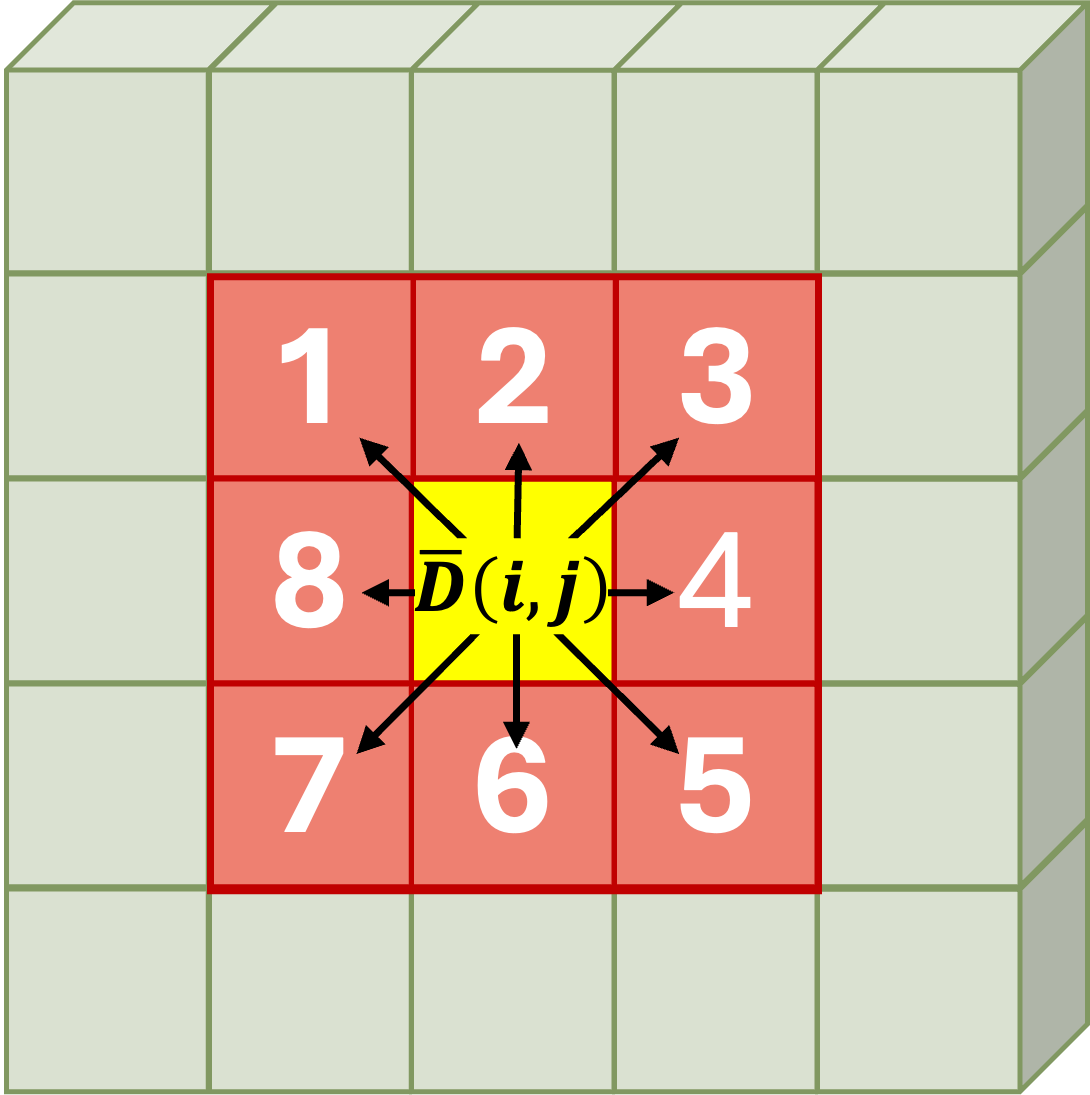}
        \caption{\textbf{$s=1$}}
    \end{subfigure}
    \hfill
    \begin{subfigure}[t]{0.28\linewidth}
        \centering
        \includegraphics[width=\linewidth]{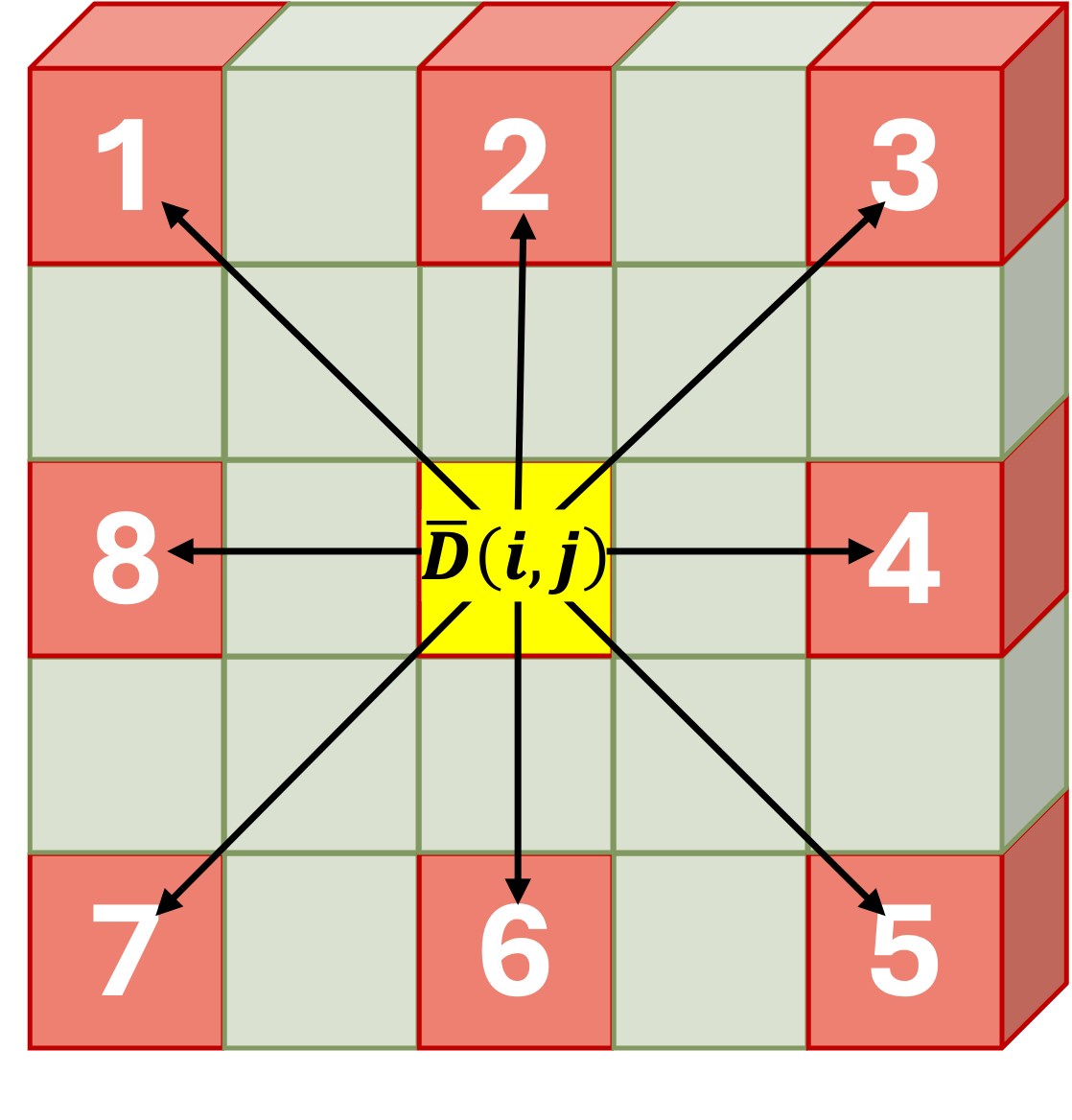}
        \caption{\textbf{$s=2$}}
    \end{subfigure}
    \hfill
    \begin{subfigure}[t]{0.28\linewidth}
        \centering
        \includegraphics[width=\linewidth]{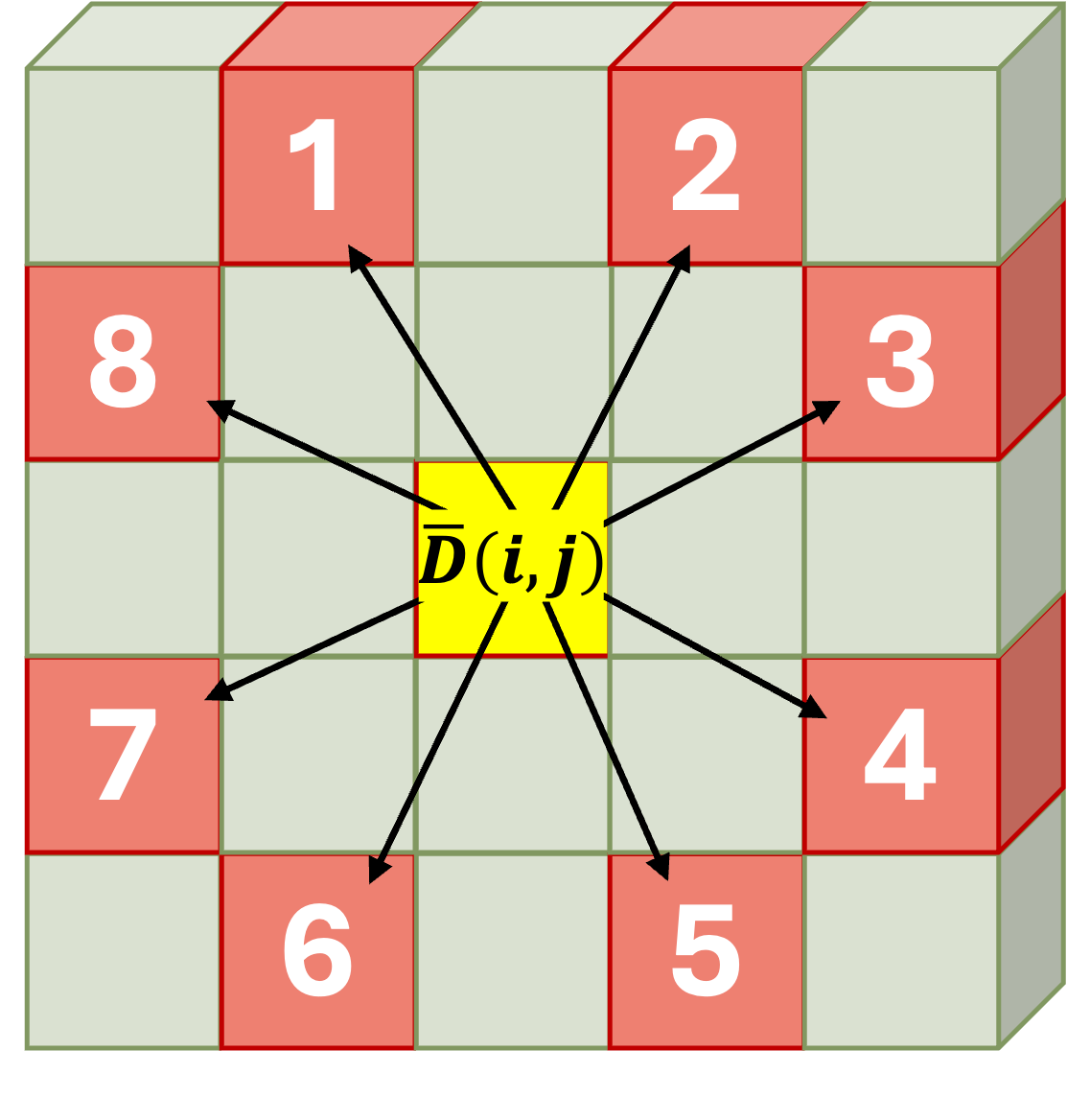}
        \caption{\textbf{$s=2'$}}
    \end{subfigure}
    \caption[Affinity Target Conversion]{
    (a) ATC samples the targets $\mathcal{A}_{s}$ from the inverted distance map $\bar{\mathcal{D}}$ by Equation~\eqref{eq:affinity}.
    (b)--(d) The sampling patterns used here: $s=1$, $s=2$, and the rotated eight-direction variant $s=2^{\prime}$.
    }
    \label{fig:affinity}
\end{figure}

\begin{figure*}[!t]
    \centering
    \begin{subfigure}[t]{0.4\textwidth}
        \centering
        \includegraphics[width=\linewidth]{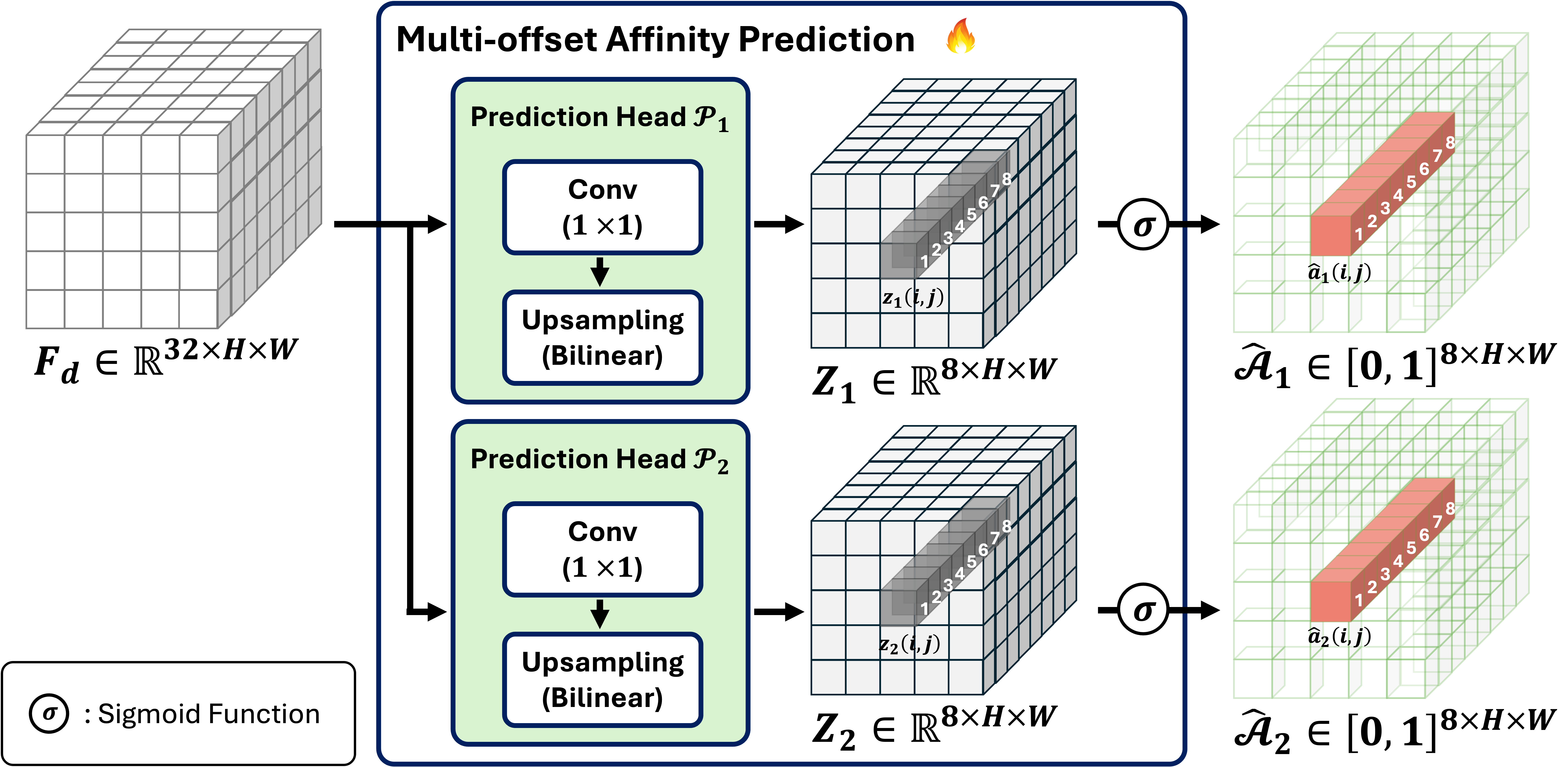}
        \par\vspace{0.5em}
        \caption{\textbf{Multi-offset Affinity Prediction}}
    \end{subfigure}
    \hfill
    \begin{subfigure}[t]{0.55\textwidth}
        \centering
        \includegraphics[width=\linewidth]{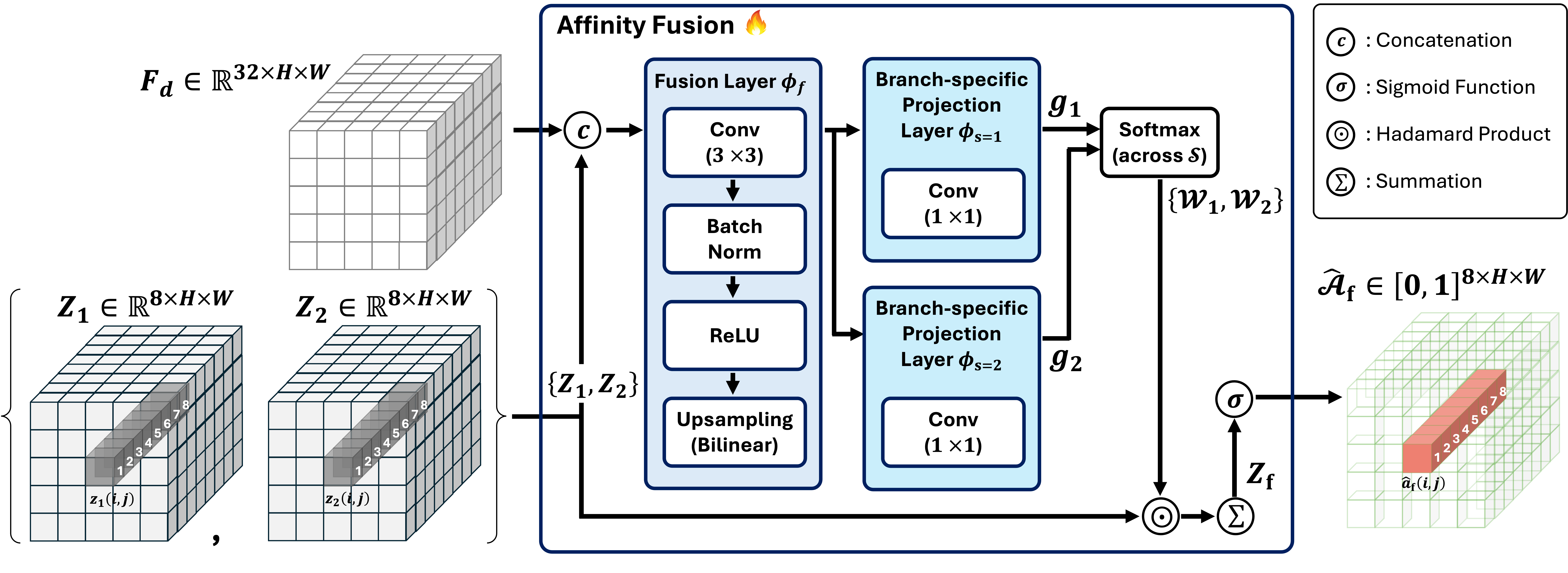}
        \par\vspace{0.5em}
        \caption{\textbf{Affinity Fusion}}
    \end{subfigure}
    \caption[Components of TMAL]{
    (a)~MAP predicts directional affinities $\{\hat{\mathcal{A}}_{s}\}_{s\in\mathcal{S}}$ from the shared decoder feature $F_\mathrm{d}$.
    (b)~$F_\mathrm{d}$ and the branch logits $\{Z_s\}_{s\in\mathcal{S}}$ are concatenated and passed through the fusion layer $\phi_{\mathrm{f}}$
    and branch-specific projection layers $\{\phi_s\}_{s\in\mathcal{S}}$ to produce per-direction spatial weights $\mathcal{W}_{s}$.
    The fused logits are obtained by summing the products of $\mathcal{W}_{s}$ and $Z_s$ across branches.
    The figure shows the main setting $\mathcal{S}=\{1,2\}$.
    }
    \label{fig:components}
\end{figure*}

\paragraph{Affinity Target Conversion.}
The IDT function replaces the binary connectivity target $\mathcal{C}$ with
the inverted distance map $\bar{\mathcal{D}}$, which is continuous.
ATC replaces it with an affinity $a^{k}_{s}(i,j)\in[0,1]$ that emphasizes thin regions near the boundary, as shown in Figure~\ref{fig:affinity}(a).

Let $\mathcal{S}$ be the set of branch indices, where each $s\in\mathcal{S}$ gives one sampling offset.
The affinity at offset $s$ is the product of the two inverted distances at the endpoints:
\begin{equation}\label{eq:affinity}
    a^{k}_{s}(i,j) = \bar{\mathcal{D}}(i,j) \cdot \bar{\mathcal{D}}(i+s\cdot o_x^k,\;j+s\cdot o_y^k),
\end{equation}
for $k=1,\ldots,8$.
Stacking the eight directions gives the continuous, direction-aware target $\mathcal{A}_{s}=\{a^{k}_{s}\}_{k=1}^{8}\in{[0,1]}^{8\times H\times W}$.
Equation~\eqref{eq:affinity} is symmetric in the two endpoints, so $a^{k}_{s}(p)=a^{\bar{k}}_{s}(q)$ at every offset $s\in\mathcal{S}$. This is equivalent to $c^{k}$ being satisfied when $s=1$ in Equation~\eqref{eq:dconn_target}.

Figure~\ref{fig:affinity}(b--d) shows the three sampling patterns used here.
In Figure~\ref{fig:affinity}(b), branch $s=1$ reads the eight neighbors of the $3\times3$ window.
Branch $s=2$ scales them onto the $5\times5$ neighborhood as shown in Figure~\ref{fig:affinity}(c).
Therefore, the branch set $\left\{\mathcal{A}_1, \mathcal{A}_2\right\}$ carries a short local step and a wider reach at the same time.
The primed branch $s=2^{\prime}$ in Figure~\ref{fig:affinity}(d) is the rotated counterpart of the axis-aligned branch $s=2$.
It keeps the same $5\times5$ window but reads the displacements $\mathcal{O}^{\prime}=\{(\pm1,\pm2),(\pm2,\pm1)\}$ instead of $\{(\pm2,0),(0,\pm2),(\pm2,\pm2)\}$.

\subsection{Feature Extraction}\label{sec:fe}
The pipeline begins with the FE module, as shown in Figure~\ref{fig:overview}(b).
FE provides the representation $F_\mathrm{d}\in\mathbb{R}^{C_d\times H\times W}$ that MAP, AF, and AuxSeg all use.
An ImageNet-pretrained ResNet-34~\cite{he2016deep} encoder $\phi_\mathrm{enc}$ extracts multi-level features from the input image $I$.
An upsampling decoder $\phi_\mathrm{dec}$ then merges them into the FE representation $F_\mathrm{d}$:
\begin{equation}\label{eq:feature_extraction}
    F_\mathrm{d}
    =
    \phi_{\mathrm{dec}}\!\left(\phi_{\mathrm{enc}}(I)\right),
\end{equation}
where $C_d=32$ is used in this study.

\subsection{Multi-offset Affinity Prediction}\label{sec:map}
MAP uses one lightweight prediction head $\mathcal{P}_{s}(\cdot)$ on the FE representation $F_\mathrm{d}$
for each branch $s\in\mathcal{S}$, as shown in Figure~\ref{fig:components}(a).
Each head applies a $1\times1$ convolution and bilinear upsampling to $F_\mathrm{d}$ and predicts an eight-channel logit map $Z_s \in \mathbb{R}^{8\times H\times W}$.
An element-wise sigmoid $\sigma(\cdot)$ then turns these logits into per-direction affinities $\hat{\mathcal{A}}_{s}\in {[0,1]}^{8\times H\times W}$:
\begin{equation}\label{eq:aff_pred}
    Z_{s}=\mathcal{P}_s(F_\mathrm{d}),
    \quad\quad
    \hat{\mathcal{A}}_{s}
    =
    \sigma\!\left(Z_{s}\right).
\end{equation}

\begin{figure*}[!t]
    \centering
    \includegraphics[width=\textwidth]{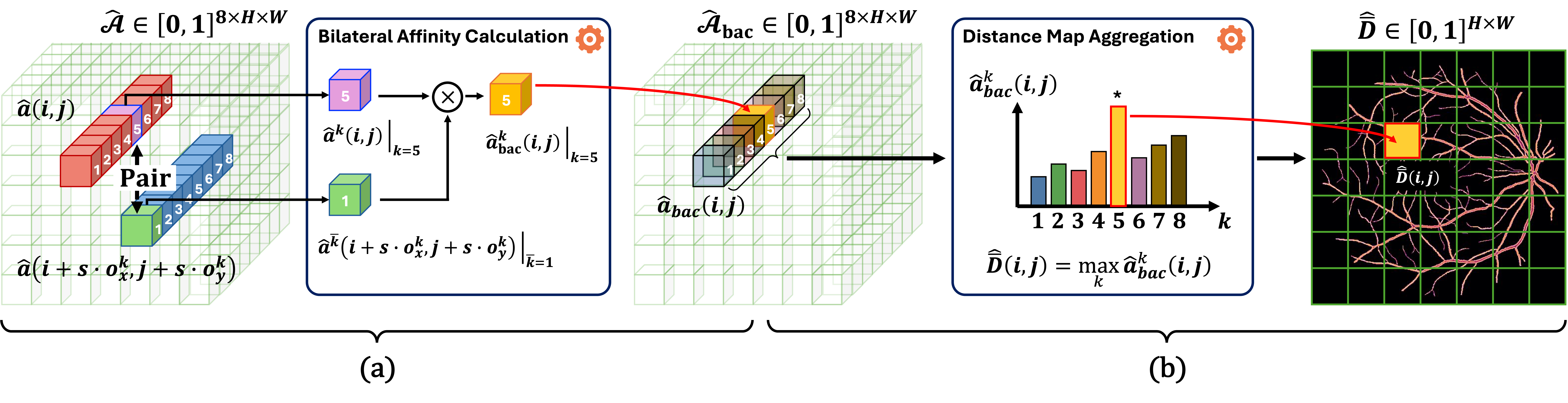}

    \caption[Distance Map Decoding]{
        DMD turns a predicted affinity map into an inverted distance map in two steps.
        (a) Bilateral Affinity Calculation multiplies the response of direction $k$ at $(i,j)$ by the response of the opposite direction $\bar{k}$ at the paired pixel.
        Stacking the eight directions gives the bilateral map $\hat{\mathcal{A}}_{\mathrm{bac}}$.
        (b) Distance Map Aggregation then takes the maximum over the eight directions and gives the predicted inverted distance map $\hat{\bar{\mathcal{D}}}$.
    }
    \label{fig:dmd}
\end{figure*}

\subsection{Affinity Fusion}\label{sec:af}
AF combines the offset branches using per-direction spatial weights, as shown in Figure~\ref{fig:components}(b).

AF first concatenates the branch logits $Z_{s}$ with the FE representation $F_\mathrm{d}$ and passes them through the fusion layer $\phi_f$, which consists of a $3\times3$ convolution, batch normalization~\cite{ioffe2015batch}, ReLU and bilinear upsampling.
A branch-specific projection layer $\phi_s$ then produces one gating logit map $g_{s}\in\mathbb{R}^{8\times H\times W}$ for each $s\in\mathcal{S}$.
A softmax over the branches turns these logits into per-direction spatial weight maps $\mathcal{W}_{s}\in{[0,1]}^{8\times H\times W}$ as
\begin{equation*}\label{eq:aff_weight}
    \mathcal{W}^{k}_{s}(i,j)
    =
    \frac{\exp\!\left(g^{k}_{s}(i,j)\right)}
    {\sum_{r\in\mathcal{S}}\exp\!\left(g^{k}_{r}(i,j)\right)},
    \quad
    \sum_{s\in\mathcal{S}}\mathcal{W}^{k}_{s}(i,j)=1 .
\end{equation*}
AF finally sums the products of each weight map $\mathcal{W}_{s}$ and logit map $Z_{s}$, and a sigmoid gives the fused affinity map $\hat{\mathcal{A}}_{\mathrm{f}}\in {[0,1]}^{8\times H\times W}$:
\begin{equation}\label{eq:aff_fusion}
    Z_{\mathrm{f}}
    =
    \sum_{s\in\mathcal{S}} \mathcal{W}_{s} \odot Z_{s},
    \quad\quad
    \hat{\mathcal{A}}_{\mathrm{f}}
    =
    \sigma\!\left(Z_{\mathrm{f}}\right),
\end{equation}
where $\odot$ denotes the Hadamard product.

We optimize all continuous targets with the Smooth L1 loss $\ell(\cdot,\cdot)$~\cite{girshick2015fast}.
The affinity predictions are supervised as
\begin{equation}\label{eq:aff_loss}
    \mathcal{L}^{(\mathrm{f})}_{\mathrm{aff}}
    =\ell\!\left(\hat{\mathcal{A}}_{\mathrm{f}},\,\mathcal{A}_{1}\right),
    \quad
    \mathcal{L}^{(s)}_{\mathrm{aff}}
    =\ell\!\left(\hat{\mathcal{A}}_{s},\,\mathcal{A}_{s}\right),
\end{equation}
where $(\mathrm{f})$ marks the fused map and $(s)$ branch $s$.
The fused map $\hat{\mathcal{A}}_{\mathrm{f}}$ from every branch set $\mathcal{S}$ is supervised by $\mathcal{L}^{(\mathrm{f})}_{\mathrm{aff}}$ to $\mathcal{A}_{1}$, because the offset-1 bilateral step of Section~\ref{sec:dmd} decodes that map.
It lets the wider offsets act as context, so that the offset-1 affinity becomes more accurate.
Each branch keeps its own offset, and $\mathcal{L}^{(s)}_{\mathrm{aff}}$ regresses $\hat{\mathcal{A}}_{s}$ against $\mathcal{A}_{s}$.

\subsection{Distance Map Decoding}\label{sec:dmd}
DMD in Figure~\ref{fig:dmd} converts a predicted affinity map $\hat{\mathcal{A}}$ into an inverted distance map $\hat{\bar{\mathcal{D}}}\in{[0,1]}^{H\times W}$ in two steps, Bilateral Affinity Calculation (BAC) and Distance Map Aggregation (DMA).
During the training phase, we apply it to the fused map $\hat{\mathcal{A}}_{\mathrm{f}}$ and each branch map $\hat{\mathcal{A}}_{s}$ for deep supervision.
The two paths give $\hat{\bar{\mathcal{D}}}^{(\mathrm{f})}$ and $\hat{\bar{\mathcal{D}}}^{(s)}$, and only $\hat{\bar{\mathcal{D}}}^{(\mathrm{f})}$ is thresholded into the final binary mask $\hat{Y}$.

\paragraph{Step 1: Bilateral Affinity Calculation.}\label{par:dmd_bac}
For every direction $k$, BAC multiplies the response at $(i,j)$ in direction $k$ by the opposite-direction $\bar{k}$ response at the paired pixel $(i + s\cdot o_x^k,\, j + s\cdot o_y^k)$, as shown in Figure~\ref{fig:dmd}(a).
The bilateral affinity map $\hat{\mathcal{A}}_{\mathrm{bac}}\in{[0,1]}^{8\times H\times W}$ is:
\begin{equation}\label{eq:bv}
    \hat{a}^{k}_{\mathrm{bac}}(i,j)
    =
    \hat{a}^{k}(i,j)
    \cdot
    \hat{a}^{\bar{k}}(i + s\cdot o_x^k,\, j + s\cdot o_y^k),
\end{equation}
where $k=1,\ldots,8$ and $\bar{k}$ is the opposite direction of $k$.
$\{\hat{a}^{k}_{\mathrm{bac}}\}_{k=1}^{8}$ is the set of all channels of the bilateral affinity map $\hat{\mathcal{A}}_{\mathrm{bac}}$.

\paragraph{Step 2: Distance Map Aggregation.}\label{par:dmd_dma}
In Figure~\ref{fig:dmd}(b), taking the maximum across directions turns $\hat{\mathcal{A}}_{\mathrm{bac}}$ into the predicted inverted distance map $\hat{\bar{\mathcal{D}}}\in[0, 1]^{H\times W}$:
\begin{equation}\label{eq:direction_collapse}
    \hat{\bar{\mathcal{D}}}(i,j)
    =
    \max_{k}\hat{a}^{k}_{\mathrm{bac}}(i,j).
\end{equation}
Collapsing a target-consistent affinity map returns $\bar{\mathcal{D}}(i,j)$ scaled by the largest inverted distance among the sampled neighbors.
That factor approaches one in the thin regions near the boundary that the target emphasizes.
Therefore, we supervise the predicted inverted distance map directly instead of inverting it:
\begin{equation}\label{eq:dist_loss}
    \mathcal{L}_{\mathrm{dist}}^{(*)}
    =
    \ell\!\left(
    \hat{\bar{\mathcal{D}}}^{(*)},\;
    \bar{\mathcal{D}}
    \right), \quad * \in \{\mathrm{f},s\}.
\end{equation}
Thresholding the predicted inverted distance map $\hat{\bar{\mathcal{D}}}^{(\mathrm{f})}$ gives the binary mask $\hat{Y}\in{\{0,1\}}^{H\times W}$:
\begin{equation}\label{eq:thresholding}
    \hat{Y}(i,j) =
    \begin{cases}
        1, & \hat{\bar{\mathcal{D}}}^{(\mathrm{f})}(i,j)>\tau, \\
        0, & \text{otherwise},
    \end{cases}
\end{equation}
where $\tau=0.5$ is used in this study.

\subsection{Auxiliary Segmentation}\label{sec:auxseg}

The AuxSeg module regularizes affinity learning with pixel-level mask supervision taken from the FE representation $F_\mathrm{d}$, as shown in Figure~\ref{fig:overview}(c).
Direct mask supervision complements the regression target and helps keep region-level accuracy.
During training the AuxSeg branch predicts a binary mask $\hat{Y}_{\mathrm{aux}}$, and a binary cross-entropy (BCE) loss supervises as
\begin{equation}\label{eq:auxseg_loss}
    \mathcal{L}_{\mathrm{aux}}
    =
    \mathrm{BCE}\!\left(\hat{Y}_{\mathrm{aux}},\,Y\right).
\end{equation}
The AuxSeg module is discarded at inference.

\begin{table*}[!t]
    \caption[Quantitative comparison on DRIVE and DeepCrack]{
        Quantitative comparison on DRIVE~\cite{staal2004ridge} and DeepCrack~\cite{liu2019deepcrack}.
        Higher is better for Dice, IoU and clDice; lower is better for $\Delta\beta_0$, $\Delta\beta_1$ and HD95.
        Methods are grouped as mask-, topology-, and connectivity-based approaches.
        Bold and underlining denote the \textbf{best} and \underline{second-best} results, respectively; the shaded row denotes TMAL.
    }
    \label{tab:main}
    \centering
    \aboverulesep=0pt
    \belowrulesep=0pt
    \footnotesize
    \renewcommand{\arraystretch}{1.2}
    \setlength{\tabcolsep}{3pt}
    \begin{tabular}{ll|l|rrrrrr|rrrrrr}
        \toprule
        \multicolumn{2}{c|}{\multirow{2}{*}{\textbf{Categories}}} & \multicolumn{1}{c|}{\multirow{2}{*}{\textbf{Methods}}}   & \multicolumn{6}{c|}{\textbf{DRIVE}~\cite{staal2004ridge}} & \multicolumn{6}{c}{\textbf{DeepCrack}~\cite{liu2019deepcrack}}                                                                                                                                                                                                                                                                                                                                                                                               \\
        \cmidrule(lr){4-9}\cmidrule(lr){10-15}
                                                                  &                                                          &                                                           & \textbf{Dice}\,$\uparrow$                                      & \textbf{IoU}\,$\uparrow$ & \textbf{clDice}\,$\uparrow$ & $\boldsymbol{\Delta\beta_0}$\,$\downarrow$ & $\boldsymbol{\Delta\beta_1}$\,$\downarrow$ & \textbf{HD95}\,$\downarrow$ & \textbf{Dice}\,$\uparrow$ & \textbf{IoU}\,$\uparrow$ & \textbf{clDice}\,$\uparrow$ & $\boldsymbol{\Delta\beta_0}$\,$\downarrow$ & $\boldsymbol{\Delta\beta_1}$\,$\downarrow$ & \textbf{HD95}\,$\downarrow$ \\
        \midrule
        \multirow{6}{*}{\textbf{(a)}}                             & \multirow{6}{*}{\textbf{Mask}}                           & U-Net~\cite{ronneberger2015u}                             & 0.7813                                                         & 0.6415                   & 0.7818                      & 446.15                                     & 6.20                                       & 9.85                        & 0.7931                    & 0.6814                   & 0.8586                      & 4.93                                       & 1.31                                       & 30.66                       \\
                                                                  &                                                          & FR-UNet~\cite{liu2022full}                                & 0.7573                                                         & 0.6097                   & 0.7716                      & 484.65                                     & 7.20                                       & 10.55                       & 0.6748                    & 0.5593                   & 0.7366                      & 5.33                                       & \underline{0.93}                           & 61.12                       \\
                                                                  &                                                          & CS-Net~\cite{mou2019cs}                                   & 0.7556                                                         & 0.6076                   & 0.7611                      & 454.65                                     & 5.85                                       & 11.41                       & 0.8051                    & 0.6945                   & 0.8721                      & 5.42                                       & 1.43                                       & 32.79                       \\ \cmidrule(lr){3-15}
                                                                  &                                                          & V-GAN~\cite{son2019towards}                               & \underline{0.8167}                                             & \underline{0.6905}       & \underline{0.8168}          & 300.65                                     & 5.50                                       & 7.44                        & 0.6249                    & 0.4933                   & 0.7021                      & 5.74                                       & 1.54                                       & 52.27                       \\
                                                                  &                                                          & SegDiff~\cite{amit2021segdiff}                            & 0.7186                                                         & 0.5617                   & 0.6998                      & 369.65                                     & 9.30                                       & 11.28                       & 0.7900                    & 0.6729                   & 0.8624                      & 5.65                                       & 1.53                                       & \textbf{14.44}              \\
                                                                  &                                                          & MedSegDiff~\cite{wu2024medsegdiff}                        & 0.7569                                                         & 0.6092                   & 0.7168                      & \underline{204.45}                         & \textbf{4.95}                              & \textbf{2.59}               & 0.7749                    & 0.6452                   & 0.8797                      & 4.62                                       & 1.70                                       & 47.89                       \\
        \midrule
        \multirow{2}{*}{\textbf{(b)}}                             & \multirow{2}{*}{\textbf{Topology}}                       & clDice~\cite{shit2021cldice}                              & 0.6561                                                         & 0.4888                   & 0.8159                      & 513.30                                     & \underline{5.20}                           & 9.08                        & 0.7673                    & 0.6642                   & 0.8258                      & \underline{4.22}                           & 2.11                                       & 48.92                       \\
                                                                  &                                                          & CAPE~\cite{esmaeilzadeh2025cape}                          & 0.7097                                                         & 0.5503                   & 0.7570                      & 307.10                                     & 6.75                                       & \underline{3.97}            & 0.8049                    & 0.6896                   & 0.8698                      & 7.08                                       & 4.27                                       & \underline{19.92}           \\
        \midrule
                                                                  &                                                          & DconnNet~\cite{yang2023directional}                       & 0.7762                                                         & 0.6346                   & 0.7835                      & 452.80                                     & 5.35                                       & 9.48                        & \underline{0.8252}        & \underline{0.7134}       & \underline{0.8927}          & 4.51                                       & 1.05                                       & 26.38                       \\
        \rowcolor{gray!15}%
        \cellcolor{white}\multirow{-2}{*}{\textbf{(c)}}           & \cellcolor{white}\multirow{-2}{*}{\textbf{Connectivity}} & \textbf{TMAL (Ours)}                                      & \textbf{0.8295}                                                & \textbf{0.7090}          & \textbf{0.8773}             & \textbf{24.50}                             & 20.95                                      & 8.48                        & \textbf{0.8439}           & \textbf{0.7427}          & \textbf{0.9074}             & \textbf{1.98}                              & \textbf{0.88}                              & 29.17                       \\
        \bottomrule
    \end{tabular}
\end{table*}

\subsection{Training Objectives}\label{sec:objectives}
All five loss terms supervise each branch, while the fused output uses only the affinity, distance, and Dice losses.
$\mathcal{L}_{\mathrm{aff}}$ optimizes all continuous targets with the Smooth L1 loss $\ell(\cdot,\cdot)$~\cite{girshick2015fast}.
$\mathcal{L}_{\mathrm{dist}}$ shapes the profile toward $\bar{\mathcal{D}}$.
$\mathcal{L}_{\mathrm{dice}}$ is the soft Dice loss~\cite{milletari2016v} between $\hat{\bar{\mathcal{D}}}^{(*)}$ and $Y$ with $*\in\{\mathrm{f},s\}$, and it counters the imbalance between foreground and background.
$\mathcal{L}_{\mathrm{bac}}$ supervises the bilateral affinity after BAC with the same offset-1 target used for local connectivity.
The edge term $\mathcal{L}_{\mathrm{edge}}$ follows the edge-decoupled supervision, and the edge set comes from the binary connectivity.

The training loss for branch $s$ is then
\begin{equation}\label{eq:branch_loss}
    \begin{aligned}
        \mathcal{L}^{(s)}
         & ={}
        \lambda_{\mathrm{aff}}\mathcal{L}^{(s)}_{\mathrm{aff}}
        +\lambda_{\mathrm{dist}}\mathcal{L}^{(s)}_{\mathrm{dist}}
        +\lambda_{\mathrm{dice}}\mathcal{L}^{(s)}_{\mathrm{dice}}
        \\ &
        +\lambda_{\mathrm{bac}}\mathcal{L}^{(s)}_{\mathrm{bac}}
        +\lambda_{\mathrm{edge}}\mathcal{L}^{(s)}_{\mathrm{edge}}.
    \end{aligned}
\end{equation}
The training loss of the fused output is
\begin{equation}\label{eq:fused_loss}
    \mathcal{L}^{(\mathrm{f})}
    =
    \lambda_{\mathrm{aff}}\mathcal{L}^{(\mathrm{f})}_{\mathrm{aff}}
    +\lambda_{\mathrm{dist}}\mathcal{L}^{(\mathrm{f})}_{\mathrm{dist}}
    +\lambda_{\mathrm{dice}}\mathcal{L}^{(\mathrm{f})}_{\mathrm{dice}},
\end{equation}
where $\lambda_{\mathrm{aff}}=0.3$, $\lambda_{\mathrm{dist}}=1.0$,
$\lambda_{\mathrm{dice}}=1.0$, $\lambda_{\mathrm{bac}}=0.2$, and $\lambda_{\mathrm{edge}}=1.0$ weight the five loss terms.

Finally, the total loss is
\begin{equation}\label{eq:total_loss}
    \mathcal{L}_{\mathrm{total}}
    =
    \lambda_{\mathrm{f}}\mathcal{L}^{(\mathrm{f})}
    +\sum_{s\in\mathcal{S}}\lambda_{s}\mathcal{L}^{(s)}
    +\lambda_{\mathrm{aux}}\mathcal{L}_{\mathrm{aux}},
\end{equation}
where $\lambda_{\mathrm{f}}=1.0$ is the weight of the fused output and $\lambda_{\mathrm{aux}}=0.5$ is the weight of the auxiliary objective.
The two branch objectives use $(\lambda_{1},\lambda_{2})=(0.5,0.3)$ for the main setting $\mathcal{S}=\{1,2\}$.
Sec. S3 in the supplementary material lists the branch weights for the ablation settings and
definition of the loss functions that are not defined in the paper.

\begin{figure*}[!t]
    \centering
    \begin{subfigure}[t]{0.84\textwidth}
        \centering
        \includegraphics[width=\linewidth]{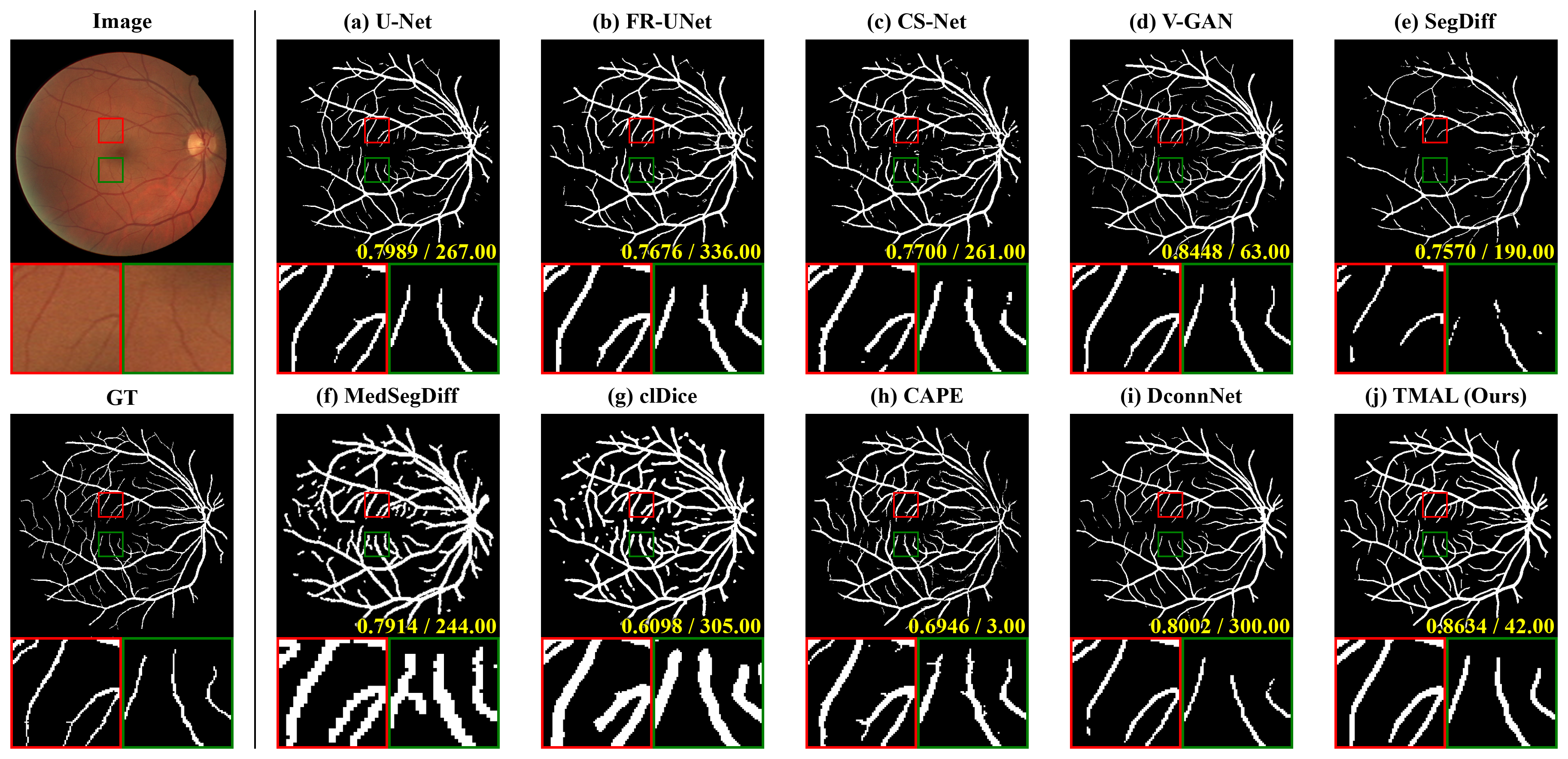}
        \caption{\textbf{DRIVE}}
    \end{subfigure}
    \\[3pt]
    \begin{subfigure}[t]{0.84\textwidth}
        \centering
        \includegraphics[width=\linewidth]{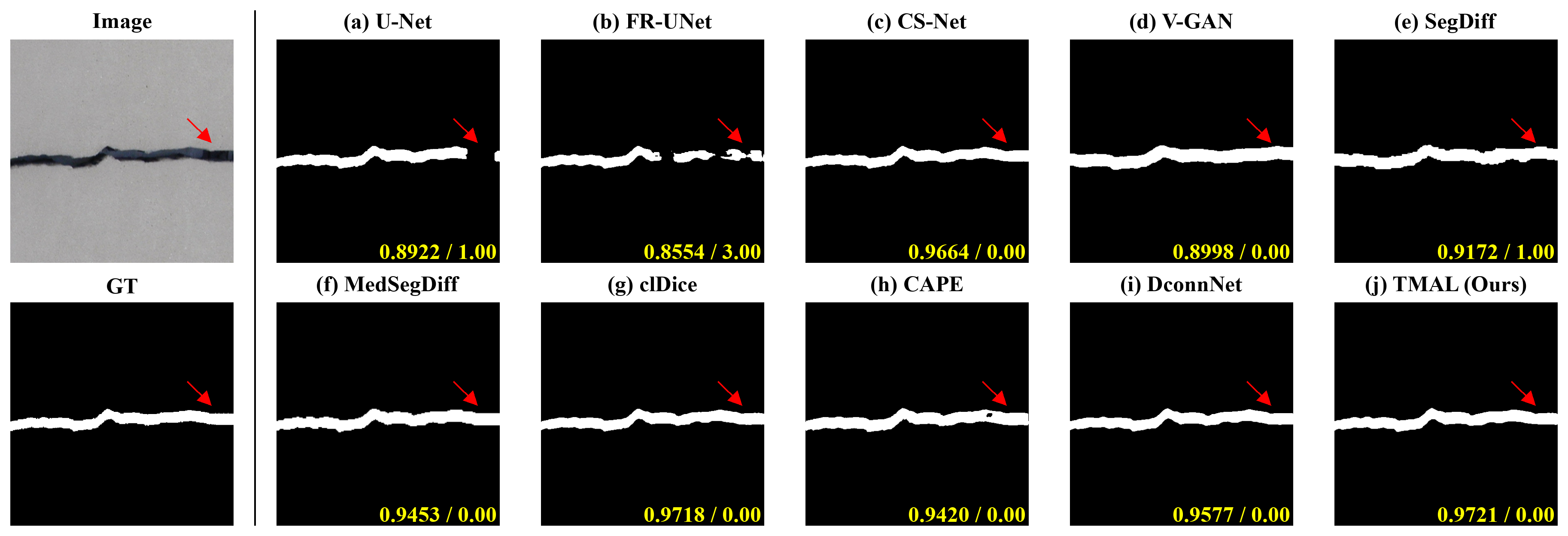}
        \caption{\textbf{DeepCrack}}
    \end{subfigure}
    \caption[Qualitative comparison on DRIVE and DeepCrack]{
        Qualitative comparison on DRIVE~\cite{staal2004ridge} and DeepCrack~\cite{liu2019deepcrack}.
        On DRIVE, red and green boxes mark the two regions enlarged beneath each panel; on DeepCrack, red arrows mark the corresponding regions.
        The highlighted regions illustrate differences in branch continuity and false-positive responses.
        Yellow values give Dice\,/\,$\Delta\beta_0$ for each example.
    }
    \label{fig:qualitative}
\end{figure*}

\section{Experiments}
\subsection{Experimental Setup}
\paragraph{Datasets.}
We evaluate TMAL on two curvilinear-structure datasets from different domains and use the official split of each.
DRIVE~\cite{staal2004ridge} provides 40 color fundus images of $565\times584$ pixels with binary vessel masks, split evenly into 20 training and 20 testing images.
DeepCrack~\cite{liu2019deepcrack} comprises 537 concrete-surface images of $544\times384$ pixels annotated for cracks, with 300 used for training and 237 for testing.

\paragraph{Implementation Details.}
FE uses the encoder and decoder of Section~\ref{sec:fe}.
Each image and its labels are resized to $512\times512$.
At inference we run the network on the same $512\times512$ input and binarize the predicted inverted distance map at $\tau=0.5$.
All metrics are computed at this resolution rather than at the native image size, and no field-of-view mask or post-processing is applied.
Sec. S1 in the supplementary material gives the optimizer, the learning rate schedule, the training length and the augmentation settings for each dataset.

\paragraph{Baseline Protocol.}
All baseline results in Table~\ref{tab:main} come from our own training runs.
Among the mask-based baselines, the encoder-decoder models U-Net~\cite{ronneberger2015u}, FR-UNet~\cite{liu2022full} and CS-Net~\cite{mou2019cs} are retrained from scratch with BCE and no architectural changes.
We retrain every other baseline with its official implementation and published configuration.
This group contains the generative V-GAN~\cite{son2019towards}, SegDiff~\cite{amit2021segdiff} and MedSegDiff~\cite{wu2024medsegdiff}, the topology-based clDice~\cite{shit2021cldice} and CAPE~\cite{esmaeilzadeh2025cape}, and the connectivity-based DconnNet~\cite{yang2023directional}.
Because every method is trained and evaluated under this common $512\times512$ protocol, the absolute values differ from those reported in the original papers.

\paragraph{Evaluation Metrics.}
We measure region overlap with Dice and IoU.
Overlap scores alone cannot capture the connectivity of thin structures, so we also report clDice~\cite{shit2021cldice}, two Betti-count errors $\Delta\beta_0$ and $\Delta\beta_1$, and the 95th percentile Hausdorff distance (HD95).
clDice measures agreement with the morphological skeleton, and the two Betti-number errors measure topology preservation.
$\Delta\beta_0$ compares the numbers of connected components and $\Delta\beta_1$ compares the numbers of loops.
HD95 measures the 95th-percentile boundary distance in pixels, taken symmetrically between the predicted and ground-truth boundaries.

\subsection{Experimental Results}
Table~\ref{tab:main} compares TMAL with mask-based, topology-based and connectivity-based methods, and Figure~\ref{fig:qualitative} shows representative predictions.
TMAL achieves the highest Dice, IoU and clDice and the lowest $\Delta\beta_0$ on both datasets.
TMAL reduces the lowest baseline $\Delta\beta_0$ by $88.0\%$ on DRIVE and by $53.1\%$ on DeepCrack, so its gains reach beyond pixel-wise accuracy to component-level continuity.

DconnNet~\cite{yang2023directional}, the closest prior method in design, uses binary connectivity targets within a fixed eight-neighborhood.
The much lower $\Delta\beta_0$ of TMAL agrees with the intended benefit of distance-aware targets and wider offsets.
In the examples of Figure~\ref{fig:qualitative},
TMAL also keeps the highlighted vessel and crack structures without an obvious break, which supports the connectivity improvements of Table~\ref{tab:main}.

\begin{table*}[!t]
    \caption[Ablation study on DRIVE]{
    Ablation study on DRIVE~\cite{staal2004ridge}.
    Each group replaces one setting, shown in parentheses.
    Removing $\mathcal{L}^{(\mathrm{f})}$ also removes AF.
    In group~(c), primed offsets denote the rotated direction set.
    Higher is better for Dice, IoU and clDice; lower is better for $\Delta\beta_0$, $\Delta\beta_1$ and HD95.
    Bold and underlining denote the \textbf{best} and \underline{second-best} results, respectively; the shaded row denotes TMAL.
    }
    \label{tab:ablation}
    \centering
    \aboverulesep=0pt
    \belowrulesep=0pt
    \footnotesize
    \renewcommand{\arraystretch}{1.2}
    \setlength{\tabcolsep}{4pt}
    \begin{tabular}{ll|cl|rrrrrr}
        \toprule
        \multicolumn{4}{c|}{\textbf{Ablation Studies}}                       & \multicolumn{6}{c}{\textbf{DRIVE}~\cite{staal2004ridge}}                                                                                                                                                                                                                                                                                                                                                                                                               \\
        \cmidrule(lr){1-4} \cmidrule(lr){5-10}
        \multicolumn{2}{c|}{\textbf{Components (Ours Setting)}}              & \multicolumn{2}{c|}{\textbf{Configurations (Other Settings)}}                                                                                                         & \textbf{Dice}\,$\uparrow$                                         & \textbf{IoU}\,$\uparrow$         & \textbf{clDice}\,$\uparrow$ & $\boldsymbol{\Delta\beta_0}$\,$\downarrow$ & $\boldsymbol{\Delta\beta_1}$\,$\downarrow$ & \textbf{HD95}\,$\downarrow$                                     \\
        \midrule
        \multirow{2}{*}{\textbf{(a)}}                                        & \multirow{2}{*}{\shortstack[l]{\textbf{Segmentation Label}                                                                                                                                                                                                                                                                                                                                                                                                             \\
        ($\bar{\mathcal{D}}$)}}                                              & (i)                                                                                                                                                                   & Binary Mask $Y$                                                   & 0.7766                           & 0.6350                      & 0.7805                                     & 480.20                                     & \textbf{5.15}               & 9.34                              \\
                                                                             &                                                                                                                                                                       & (ii)                                                              & Distance Map $\mathcal{D}$       & 0.8268                      & 0.7051                                     & \underline{0.8760}                         & \underline{27.95}           & 21.15             & 8.52          \\
        \midrule
        \multirow{3}{*}{\textbf{(b)}}                                        & \multirow{3}{*}{\shortstack[l]{\textbf{Objective Terms}                                                                                                                                                                                                                                                                                                                                                                                                                \\
        (w/ $\mathcal{L}^{(\mathrm{f})}$ and $\mathcal{L}_{\mathrm{aux}}$)}} & (i)                                                                                                                                                                   & w/o $\mathcal{L}^{(\mathrm{f})}$ and $\mathcal{L}_{\mathrm{aux}}$ & 0.8242                           & 0.7013                      & 0.8745                                     & 67.65                                      & 21.95                       & 8.42                              \\
                                                                             &                                                                                                                                                                       & (ii)                                                              & w/o $\mathcal{L}^{(\mathrm{f})}$ & 0.8269                      & 0.7052                                     & 0.8752                                     & 68.50                       & 21.75             & 8.54          \\
                                                                             &                                                                                                                                                                       & (iii)                                                             & w/o $\mathcal{L}_{\mathrm{aux}}$ & 0.8243                      & 0.7015                                     & 0.8720                                     & 31.35                       & 21.20             & 8.45          \\
        \midrule
        \multirow{6}{*}{\textbf{(c)}}                                        & \multirow{6}{*}{\shortstack[l]{\textbf{Branch Set}                                                                                                                                                                                                                                                                                                                                                                                                                     \\
        ($\mathcal{S}=\{1,2\}$)}}                                            & (i)                                                                                                                                                                   & $\mathcal{S}=\{1\}$                                               & 0.8264                           & 0.7045                      & 0.8759                                     & 30.10                                      & 20.95                       & \underline{8.36}                  \\
                                                                             &                                                                                                                                                                       & (ii)                                                              & $\mathcal{S}=\{2\}$              & 0.8184                      & 0.6928                                     & 0.8648                                     & 44.50                       & \underline{17.90} & 8.55          \\
        \cmidrule(lr){3-10}
                                                                             &                                                                                                                                                                       & (iii)                                                             & $\mathcal{S}=\{1,2,3\}$          & \underline{0.8277}          & \underline{0.7064}                         & 0.8748                                     & 29.05                       & 20.70             & 8.70          \\
                                                                             &                                                                                                                                                                       & (iv)                                                              & $\mathcal{S}=\{1,2,3,4\}$        & 0.8266                      & 0.7048                                     & 0.8742                                     & 33.50                       & 21.20             & 8.50          \\
        \cmidrule(lr){3-10}
                                                                             &                                                                                                                                                                       & (v)                                                               & $\mathcal{S}=\{1,2^{\prime}\}$   & 0.8265                      & 0.7046                                     & 0.8745                                     & 35.95                       & 21.25             & 8.37          \\
                                                                             &                                                                                                                                                                       & (vi)                                                              & $\mathcal{S}=\{1,2,2^{\prime}\}$ & 0.8270                      & 0.7053                                     & 0.8756                                     & 34.10                       & 20.75             & \textbf{8.16} \\
        \midrule
        \rowcolor{gray!15}
        \textbf{(d)}                                                         & \multicolumn{3}{l|}{\textbf{TMAL (Ours)}\quad(Setting: $\bar{\mathcal{D}}$, w/ $\mathcal{L}^{(\mathrm{f})}$ and $\mathcal{L}_{\mathrm{aux}}$, $\mathcal{S}=\{1,2\}$)} & \textbf{0.8295}                                                   & \textbf{0.7090}                  & \textbf{0.8773}             & \textbf{24.50}                             & 20.95                                      & 8.48                                                            \\
        \bottomrule
    \end{tabular}
\end{table*}

\subsection{Ablation Studies}\label{sec:ablation}
Table~\ref{tab:ablation} ablates TMAL on DRIVE along three axes: the segmentation label, the objective terms and the branch set. Figures of each ablation study is in Sec. S5 of the supplementary material.

\paragraph{Segmentation Label.}
Table~\ref{tab:ablation}(a) compares three segmentation labels, the binary mask $Y$, the standard distance map $\mathcal{D}$ and the inverted distance map $\bar{\mathcal{D}}$.
Replacing the soft affinity target with the binary mask $Y$ in Table~\ref{tab:ablation}(a)(i) reduces Dice, IoU and clDice, and increases $\Delta\beta_0$ by more than an order of magnitude.
However, the binary mask attains the lowest $\Delta\beta_1$ of the study, namely $5.15$ against $20.95$ for the inverted distance map $\bar{\mathcal{D}}$.
A binary mask leaves thin branches broken, whereas a soft affinity target reconnects them and can also close a loop the ground truth does not contain.
The standard distance map $\mathcal{D}$ in Table~\ref{tab:ablation}(a)(ii) is worse on every metric than $\bar{\mathcal{D}}$ in~(d), which supports emphasizing thin regions near the boundary instead of structure centerlines.

\paragraph{Objective Terms.}
Table~\ref{tab:ablation}(b) ablates the fused objective $\mathcal{L}^{(\mathrm{f})}$ and the auxiliary objective $\mathcal{L}_{\mathrm{aux}}$.
The full TMAL model in~(d) attains a Betti-0 error of $\Delta\beta_0=24.50$.
Removing both terms in~(b)(i) increases this error to $67.65$, and removing $\mathcal{L}^{(\mathrm{f})}$ alone in~(b)(ii) yields a comparable $68.50$; the overlap scores and $\Delta\beta_1$ degrade in the same way.
The degradation is therefore driven almost entirely by $\mathcal{L}^{(\mathrm{f})}$, although this comparison should be read with care, since removing $\mathcal{L}^{(\mathrm{f})}$ also leaves the fused path unsupervised and~(b)(ii) thus cannot separate the absence of the loss from an untrained fusion branch.
Omitting $\mathcal{L}_{\mathrm{aux}}$ alone in~(b)(iii) is far less harmful, raising $\Delta\beta_0$ to only $31.35$, but it still lowers the overlap scores and increases both Betti-number errors relative to~(d), indicating that the auxiliary supervision provides useful regularization for affinity learning.

\paragraph{Branch Set.}
Table~\ref{tab:ablation}(c) varies the branch set $\mathcal{S}$ in two respects: the number of sampling offsets, and whether the direction set at offset~2 is rotated.
A single offset yields a higher $\Delta\beta_0$ than the proposed pair, both for the short step $\mathcal{A}_{1}$ in~(c)(i) and for the wider step $\mathcal{A}_{2}$ in~(c)(ii).
Enlarging the set beyond two offsets trades topological accuracy for overlap accuracy: $\mathcal{S}=\{1,2,3\}$ in~(c)(iii) attains the best Dice and IoU among the ablated variants, yet it and $\mathcal{S}=\{1,2,3,4\}$ in~(c)(iv) both raise $\Delta\beta_0$ above the proposed setting $\mathcal{S}=\{1,2\}$.
To probe non-axis-aligned directions, we introduce the rotated branch $s=2^{\prime}$ of Figure~\ref{fig:affinity}(d) in two ways: in~(c)(v) it replaces $s=2$, giving $\mathcal{S}=\{1,2^{\prime}\}$, and in~(c)(vi) it is added to the proposed setting, giving $\mathcal{S}=\{1,2,2^{\prime}\}$.
Under the same setting, neither variant improves the overlap scores or $\Delta\beta_0$.
However, variant~(c)(vi) does attain the best HD95 in Table~\ref{tab:ablation}, consistent with the uniform sampling radius of $\mathcal{O}^{\prime}$ aiding boundary localization.
Therefore, we adopt $\mathcal{S}=\{1,2\}$ for TMAL.

\section{Conclusion}
We presented TMAL, which replaces binary connectivity targets with soft directional affinities built from the inverted distance map, so the supervision follows local thickness and marks the structures that break most easily.
TMAL predicts these affinities at several sampling offsets and combines them with adaptive weights.
On DRIVE and DeepCrack it reduces the connected-component error $\Delta\beta_0$ and also improves the overlap and skeleton-based metrics.

\paragraph{Limitations.}
The affinity target constrains pairwise connectivity, so it does not directly optimize loop structure or the worst-case boundary distance.
$\Delta\beta_1$ rises on DRIVE and HD95 does not reach the best value.
The sampling offsets are also fixed in advance and shared across the image.

\paragraph{Future Work.}
An affinity target that also constrains loops and boundaries would turn these properties into a controllable trade-off.
The sampling offsets could likewise be learned.

\section*{Acknowledgments}
This work was supported by the IITP(Institute of Information \& Coummunications Technology Planning \& Evaluation)-ITRC(Information Technology Research Center) grant funded by the Korea government(Ministry of Science and ICT)(IITP-2026-RS-2020-II201602),
by the National Research Foundation of Korea(NRF) grant funded by the Korea government(MSIT) (RS-2025-24533373),
by the Cyber Investigation Support Technology Development Program(No.RS-2025-02304983) of the Korea Institute of Police Technology (KIPoT) funded by the Korean National Police Agency
and by the G-LAMP Program of the National Research Foundation of Korea (NRF) grant funded by the Ministry of Education (No. RS-2025-25441317).

\clearpage
\setcounter{section}{0}
\setcounter{figure}{0}
\setcounter{table}{0}
\setcounter{equation}{0}
\renewcommand{\thetable}{S\arabic{table}}
\renewcommand{\thefigure}{S\arabic{figure}}
\renewcommand{\theequation}{S\arabic{equation}}
\renewcommand{\thesection}{S\arabic{section}}
\renewcommand{\theHtable}{S\arabic{table}}
\renewcommand{\theHfigure}{S\arabic{figure}}
\renewcommand{\theHequation}{S\arabic{equation}}
\renewcommand{\theHsection}{S\arabic{section}}
\makeatletter
\setlength{\@fptop}{0pt}
\makeatother

\maketitlesupplementary
\vspace{-1.25em}

This document provides additional material for the main paper.
It includes implementation details in Section~\ref{sec:sup_impl}, additional explanation about the affinity target $\mathcal{A}$ in Section~\ref{sec:sup_target},
the full loss definitions in Section~\ref{sec:sup_loss}, metric definitions in Section~\ref{sec:sup_metrics},
and additional analysis and qualitative results in Section~\ref{sec:sup_qual}.

Following the main paper, $\mathcal{S}$ is the set of branch indices and each $s\in\mathcal{S}$ gives one sampling offset,
$F_\mathrm{d}$ denotes the decoder feature produced by Feature Extraction (FE),
and $\bar{\mathcal{D}}$ denotes the normalized inverted distance map.
The superscript $(\mathrm{f})$ denotes a term for the fused map and $(s)$ the corresponding term for branch $s$.

\section{Implementation Details}
\label{sec:sup_impl}

\paragraph{Preprocessing}
We resize the images, binary masks, and distance maps to $512\times512$ and normalize image intensities to $[-1.6,\,1.6]$.
Training augmentation is HSV jitter with hue, saturation and value shifts of at most $30$, $5$ and $15$ levels, a random shift, scale and aspect-ratio jitter of at most $10\%$,
random horizontal and vertical flips, and random $90^{\circ}$ rotations.
We do not use test-time augmentation.

\paragraph{Optimization}
We train TMAL on one NVIDIA RTX PRO 6000 GPU\@, with Adam on DRIVE and AdamW on DeepCrack.
The baselines use their published optimizer settings.
We use a separate schedule for each dataset, as shown in Table~\ref{tab:sup_optim}.
DeepCrack has 300 training images, whereas DRIVE has 20, so we train DeepCrack for fewer epochs with a lower learning rate.
We fix the random seed and enable deterministic kernels.
We hold out no validation split, so no checkpoint selection or early stopping takes place and
every reported number comes from the last-epoch model.

\begin{table}[htbp]
    \caption{Optimization settings of the reported runs.}
    \label{tab:sup_optim}
    \centering
    \setlength{\tabcolsep}{4pt}
    \begin{tabular}{l|cc}
        \toprule
                      & DRIVE~\cite{S-staal2004ridge} & DeepCrack~\cite{S-liu2019deepcrack} \\
        \midrule
        Epochs        & 500                           & 100                                 \\
        Batch size    & 4                             & 8                                   \\
        Optimizer     & Adam                          & AdamW                               \\
        Learning rate & $3.8\times10^{-3}$            & $1\times10^{-4}$                    \\
        Schedule      & polynomial ($0.9$)            & cosine annealing                    \\
        Weight decay  & 0                             & $1\times10^{-4}$                    \\
        \bottomrule
    \end{tabular}
\end{table}

\paragraph{Inference}
At inference we run the network once on the same $512\times512$ input and
binarize the predicted inverted distance map at $\tau=0.5$. We discard AuxSeg, so only the
affinity path produces the reported binary mask.

\paragraph{Auxiliary Segmentation}
AuxSeg is a training-only one-channel mask head.
In the main paper, we directly apply the auxiliary head $\phi_{\mathrm{aux}}$ to the decoder feature $F_\mathrm{d}$:
\begin{equation}\label{eq:sup_auxseg_head}
    \begin{aligned}
        M_{\mathrm{aux}}
         & =
        \phi_{\mathrm{aux}}(F_\mathrm{d}),\qquad
        \hat{Y}_{\mathrm{aux}}=\sigma(M_{\mathrm{aux}}).
    \end{aligned}
\end{equation}
Here $\phi_{\mathrm{aux}}$ denotes a shallow convolutional head,
implemented as a $3\times3$ convolution followed by batch normalization and ReLU, and then a $1\times1$ convolution:
$\phi_{\mathrm{aux}}(\cdot)=\operatorname{Conv}_{1\times1}(\operatorname{ReLU}(\operatorname{BN}(\operatorname{Conv}_{3\times3}(\cdot))))$.
The decoder feature has $C_d=32$ channels.
The auxiliary loss is the binary cross-entropy of Equation~(13) of the main paper,
\begin{equation}\label{eq:sup_auxseg_loss}
    \mathcal{L}_{\mathrm{aux}}
    =
    \mathrm{BCE}\!\left(\hat{Y}_{\mathrm{aux}},\,Y\right),
\end{equation}
which we evaluate as $\operatorname{BCEWithLogits}(M_{\mathrm{aux}},Y)$ on the logits for numerical stability.
AuxSeg is discarded at inference, so it changes only the training objective and not the deployed affinity-decoding path.

\paragraph{Choice of Fusion Mode in AF}
Table~\ref{tab:sup_fusion_mode} compares the fusion choices under the reported inverted-distance setting.

The proposed configuration uses Decoder-guided Directional Offset Fusion (DG-DOF), which fuses branch logits rather than affinity probabilities.
It concatenates the decoder feature $F_\mathrm{d}$ with all branch logits $\{Z_s\}_{s\in\mathcal{S}}$, applies the shared fusion layer $\phi_{\mathrm{f}}$, and predicts branch-specific gating logits through $\phi_s$.
A softmax across branches gives per-pixel, per-direction weights $\mathcal{W}_s$, so each fused directional logit is an adaptive convex combination over offsets.
This choice keeps the offset-specific branches supervised separately, while the fused map, supervised by the offset-1 target, uses wider-offset evidence as context for the final bilateral decoding.
For the two-branch setting below, $Z_1$ and $Z_2$ denote the branch logits at offsets 1 and 2, respectively.

No Fusion (NF) in Equation~\ref{eq:sup_fusion_nf} decodes $Z_1$ while training both branches.
Decoder-guided Residual (DGR) in Equation~\ref{eq:sup_fusion_dgr} adds to $Z_1$ a residual from $F_\mathrm{d}$, $Z_1$, and $Z_2$.
DGR outperformed NF; logit fusion improves numerical stability.
Directional Offset Fusion (DOF) in Equation~\ref{eq:sup_fusion_dof} convexly combines $Z_1$ and $Z_2$.
DG-DOF in Equation~\ref{eq:sup_fusion_dgdof} conditions its weights on $F_\mathrm{d}$ and serves as Affinity Fusion (AF).

The candidate fusion modes are
\begin{equation}\label{eq:sup_fusion_nf}
    \text{NF:}\quad Z_\mathrm{f}=Z_1 .
\end{equation}
\begin{equation}\label{eq:sup_fusion_dgr}
    \begin{aligned}
        \text{DGR:}\quad
        Z_\mathrm{f}
         & = Z_1+\beta\odot r,\quad r=\phi_r(h_\mathrm{dgr}), \\
        \beta
         & = \sigma\!\left(\phi_\beta(h_\mathrm{dgr})\right), \\
        h_\mathrm{dgr}
         & = \phi_\mathrm{dgr}([F_\mathrm{d},Z_1,Z_2]).
    \end{aligned}
\end{equation}
\begin{equation}\label{eq:sup_fusion_dof}
    \begin{aligned}
        \text{DOF:}\quad
        Z_\mathrm{f}
         & = \sum_{s\in\mathcal{S}}\mathcal{W}_s\odot Z_s,                    \\
        \mathcal{W}_s
         & = \operatorname{softmax}_{s}\!\left(\phi_s(h_\mathrm{dof})\right), \\
        h_\mathrm{dof}
         & = \phi_\mathrm{dof}([Z_1,Z_2]).
    \end{aligned}
\end{equation}
\begin{equation}\label{eq:sup_fusion_dgdof}
    \begin{aligned}
        \text{DG-DOF:}\quad
        Z_\mathrm{f}
         & = \sum_{s\in\mathcal{S}}\mathcal{W}_s\odot Z_s,                  \\
        \mathcal{W}_s
         & = \operatorname{softmax}_{s}\!\left(\phi_s(h_\mathrm{f})\right), \\
        h_\mathrm{f}
         & = \phi_{f}([F_\mathrm{d},Z_1,Z_2]).
    \end{aligned}
\end{equation}

Each mode has its own fusion layer.
For DG-DOF, $\phi_f$ is the fusion layer and $\phi_s$ is the branch-specific projection layer of the main paper.
The notation $[\cdot]$ denotes channel concatenation, and all weights are spatial and direction-specific.
Table~\ref{tab:sup_fusion_mode} reports these three multi-offset variants together with the no-fusion baseline.

\begin{table}[!t]
    \caption{Fusion-mode ablation under the inverted-distance setting.
        All rows use the standard eight directions, Smooth L1 supervision, AuxSeg weight $0.5$, and Distance Map Aggregation by the maximum over the eight directions.
        All rows use $\mathcal{S}=\{1,2\}$ with branch loss weights $(\lambda_{1},\lambda_{2})=(0.5,0.3)$.
        The no-fusion row keeps both branches but has no AF module, so the offset-1 branch logits are decoded directly as in Equation~\eqref{eq:sup_fusion_nf}.
        Higher is better for Dice, IoU and clDice; lower is better for $\Delta\beta_0$, $\Delta\beta_1$ and HD95.
        Bold denotes the \textbf{best} result in each column; the shaded row is the proposed configuration.}
    \label{tab:sup_fusion_mode}
    \centering
    \aboverulesep=0pt
    \belowrulesep=0pt
    \footnotesize
    \renewcommand{\arraystretch}{1.12}
    \setlength{\tabcolsep}{2pt}
    \fitwidth[\columnwidth]{%
        \begin{tabular}{l|rrrrrr}
            \toprule
            \multirow{2}{*}{\textbf{Fusion mode}} & \multicolumn{6}{c}{\textbf{DRIVE}~\cite{S-staal2004ridge}}                                                                                                                                                                                  \\
            \cmidrule(lr){2-7}
                                                  & \textbf{Dice}\,$\uparrow$                                  & \textbf{IoU}\,$\uparrow$ & \textbf{clDice}\,$\uparrow$ & $\boldsymbol{\Delta\beta_0}$\,$\downarrow$ & $\boldsymbol{\Delta\beta_1}$\,$\downarrow$ & \textbf{HD95}\,$\downarrow$ \\
            \midrule
            NF                                    & 0.8269                                                     & 0.7052                   & 0.8752                      & 68.50                                      & 21.75                                      & 8.54                        \\
            DGR                                   & 0.8285                                                     & 0.7076                   & 0.8760                      & 28.85                                      & 21.10                                      & \textbf{8.37}               \\
            DOF                                   & 0.8248                                                     & 0.7021                   & 0.8743                      & 25.10                                      & \textbf{20.40}                             & 8.57                        \\
            \rowcolor{gray!15}
            \textbf{DG-DOF}                       & \textbf{0.8295}                                            & \textbf{0.7090}          & \textbf{0.8773}             & \textbf{24.50}                             & 20.95                                      & 8.48                        \\
            \bottomrule
        \end{tabular}}
\end{table}

\section{Affinity Target Details}
\label{sec:sup_target}

This section collects properties of the affinity target that the main paper
states only in summary.

\paragraph{Degenerate Cases}
We construct the normalized inverted distance map as
\begin{equation}\label{eq:sup_inv_norm}
    \bar{\mathcal{D}}(i,j)
    =
    \begin{cases}
        \dfrac{d_{\max}-\mathcal{D}(i,j)}{d_{\max}-d_{\min}}, & (i,j)\in\Omega_1, \\
        0,                                                    & (i,j)\in\Omega_0.
    \end{cases}
\end{equation}
If $d_{\max}=d_{\min}$, the denominator vanishes and the normalization can cause a division-by-zero error.
However, none of the samples in DRIVE or DeepCrack exhibited this degenerate case.

\paragraph{Rotated Direction Set}
The primed branch $2^{\prime}$ replaces the axis-aligned offsets $\{(\pm2,0),(0,\pm2),(\pm2,\pm2)\}$ at offset two with
$\{(\pm1,\pm2),(\pm2,\pm1)\}$.
Both sets have eight elements inside the same $5\times5$ window and both are closed under negation,
so the opposite-direction indexing of the main paper, and with it the shared target of a direction pair, applies unchanged.
The rotated set places all eight samples at radius $\sqrt5$, whereas the axis-aligned set mixes radii $2$ and $2\sqrt2$.

\section{Loss Definitions}
\label{sec:sup_loss}

\paragraph{Smooth L1}
We use the Smooth L1 loss $\ell(\cdot, \cdot)$~\cite{S-girshick2015fast} for all continuous affinity and distance targets.
The loss is applied element-wise and averaged over the map:
\begin{equation}\label{eq:sup_smoothl1}
    \ell(x,y)
    =
    \begin{cases}
        \dfrac{0.5\,(x-y)^2}{\delta}, & |x-y| < \delta,   \\[4pt]
        |x-y| - \tfrac{1}{2}\delta,   & \text{otherwise},
    \end{cases}
\end{equation}
with $\delta=1.0$ in all experiments.

\paragraph{Bilateral Consistency Term}
After Bilateral Affinity Calculation, we compare
$\hat{\mathcal{A}}^{(s)}_{\mathrm{bac}}=\{\hat{a}^{k,(s)}_{\mathrm{bac}}\}_{k=1}^{8}$
of branch $s$ with the offset-1 affinity target used for local connectivity:
\begin{equation}\label{eq:sup_bac}
    \mathcal{L}^{(s)}_{\mathrm{bac}}
    =
    \ell\!\left(\hat{\mathcal{A}}^{(s)}_{\mathrm{bac}},\,\mathcal{A}_{1}\right).
\end{equation}
The offset-1 affinity target is symmetric in the two endpoints, so both directions have the same target value.
Following BiconNet~\cite{S-yang2022biconnet}, we reuse this target for both the pre-BAC directional fused map and every branch's bilateral map.
This reuse is a design choice, not an algebraic identity: if a pre-BAC directional term were satisfied exactly,
the bilateral product would equal the elementwise square of its prediction.
The bilateral term therefore couples the two halves of each direction pair through their product rather than acting as a constraint that can be met exactly.
We apply it to the branches only.

\paragraph{Dice Term}
Each decoded inverted distance map also carries a soft Dice loss~\cite{S-milletari2016v} against the ground-truth mask $Y$,
\begin{equation}\label{eq:sup_dice}
    \mathcal{L}^{(*)}_{\mathrm{dice}}
    =
    1-
    \frac{2\sum_{i,j}\hat{\bar{\mathcal{D}}}^{(*)}(i,j)\,y(i,j)+\varepsilon}
    {\sum_{i,j}\hat{\bar{\mathcal{D}}}^{(*)}(i,j)+\sum_{i,j}y(i,j)+\varepsilon},
    \quad *\in\{\mathrm{f},s\},
\end{equation}
where $\varepsilon=10^{-4}$.
This loss provides region-level supervision for the affinity path.
Unlike the Size Density Loss of DconnNet~\cite{S-yang2023directional}, it does not weight classes by size because both benchmarks contain one foreground class.

\paragraph{Edge-decoupled Term}
Each branch loss also includes the edge-decoupled term of BiconNet~\cite{S-yang2022biconnet}.
Let $c^k$ denote the binary connectivity target from the main paper.
Because our affinity targets are continuous, we construct $\mathcal{C}=\{c^k\}_{k=1}^{8}$ from the ground-truth mask $Y$ rather than from the distance-aware affinity target.
The binary edge target is calculated from the connectivity target as
\begin{equation}\label{eq:sup_edge_target}
    \begin{aligned}
        e(i,j)
         & =
        \mathbf{1}\!\left[
            0 < \sum_{k=1}^{8} c^k(i,j) < 8
            \right],
        \\
        \mathcal{E}
         & =
        \{(i,j)\mid e(i,j)=1\}.
    \end{aligned}
\end{equation}
Thus, an edge pixel is a foreground pixel whose connectivity vector contains both zeros and ones.
Following the edge-decoupled formulation of BiconNet~\cite{S-yang2022biconnet}, which DconnNet~\cite{S-yang2023directional} adopts as part of its Bicon loss, the per-pixel edge term for branch $s$ is
\begin{equation}\label{eq:sup_edgemap}
    \begin{aligned}
         & \ell_{\mathrm{BCE}}\!\left(
        1-\min_{k}\hat{a}^{k,(s)}_{\mathrm{bac}}(i,j),\,
        y(i,j)
        \right)
        \\
         & \qquad={}
        \ell_{\mathrm{BCE}}\!\left(
        \min_{k}\hat{a}^{k,(s)}_{\mathrm{bac}}(i,j),\,0
        \right),
        \quad (i,j)\in\mathcal{E},
    \end{aligned}
\end{equation}
where $y(i,j)=1$ for every $(i,j)\in\mathcal{E}$.
This drives the smallest bilateral response, corresponding to a disconnected direction, toward zero.
We sum this term over the edge pixels and normalize it by the mass of their minimum bilateral responses,
\begin{equation}\label{eq:sup_edge}
    \mathcal{L}^{(s)}_{\mathrm{edge}}
    =
    \frac{
    -\sum_{(i,j)\in\mathcal{E}}
    \log\!\left(1-\min_{k}\hat{a}^{k,(s)}_{\mathrm{bac}}(i,j)\right)
    }
    {
    \max\!\left(
    \sum_{(i,j)\in\mathcal{E}}
    \min_{k}\hat{a}^{k,(s)}_{\mathrm{bac}}(i,j),
    \,10^{-8}
    \right)
    }.
\end{equation}

\paragraph{Total Objective}
The total objective groups losses by decoded map rather than by term type.
Each branch $s$ collects all five terms at its own sampling offset:
\begin{equation}\label{eq:sup_group}
    \begin{aligned}
        \mathcal{L}^{(s)}
         & ={}
        \lambda_{\mathrm{aff}}\mathcal{L}^{(s)}_{\mathrm{aff}}
        +\lambda_{\mathrm{dist}}\mathcal{L}^{(s)}_{\mathrm{dist}}
        \\&
        +\lambda_{\mathrm{dice}}\mathcal{L}^{(s)}_{\mathrm{dice}}
        +\lambda_{\mathrm{bac}}\mathcal{L}^{(s)}_{\mathrm{bac}}
        +\lambda_{\mathrm{edge}}\mathcal{L}^{(s)}_{\mathrm{edge}}.
    \end{aligned}
\end{equation}
The fused map contributes only three of them. Its bilateral and edge terms are not part of the objective:
\begin{equation}\label{eq:sup_fused}
    \mathcal{L}^{(\mathrm{f})}
    =
    \lambda_{\mathrm{aff}}\mathcal{L}^{(\mathrm{f})}_{\mathrm{aff}}
    +\lambda_{\mathrm{dist}}\mathcal{L}^{(\mathrm{f})}_{\mathrm{dist}}
    +\lambda_{\mathrm{dice}}\mathcal{L}^{(\mathrm{f})}_{\mathrm{dice}}.
\end{equation}
The total objective adds the auxiliary branch:
\begin{equation}\label{eq:sup_total}
    \mathcal{L}_{\mathrm{total}}
    =
    \lambda_{\mathrm{f}}\mathcal{L}^{(\mathrm{f})}
    +\sum_{s\in\mathcal{S}}\lambda_{s}\mathcal{L}^{(s)}
    +\lambda_{\mathrm{aux}}\mathcal{L}_{\mathrm{aux}}.
\end{equation}
We set $\lambda_{\mathrm{aff}}=0.3$,
$\lambda_{\mathrm{dist}}=1.0$,
$\lambda_{\mathrm{dice}}=1.0$,
$\lambda_{\mathrm{bac}}=0.2$,
$\lambda_{\mathrm{edge}}=1.0$,
$(\lambda_{1},\lambda_{2})=(0.5, 0.3)$, $\lambda_{\mathrm{f}}=1.0$, and $\lambda_{\mathrm{aux}}=0.5$.
For the branch sets with more than two offsets we additionally set $(\lambda_{3},\lambda_{4})=(0.3, 0.2)$.

\section{Metric Definitions}
\label{sec:sup_metrics}

\begin{figure*}[!t]
    \centering
    \captionsetup[subfigure]{skip=1pt,belowskip=-4pt}
    \begin{subfigure}[t]{0.23\textwidth}
        \centering
        \includegraphics[width=0.948\linewidth]{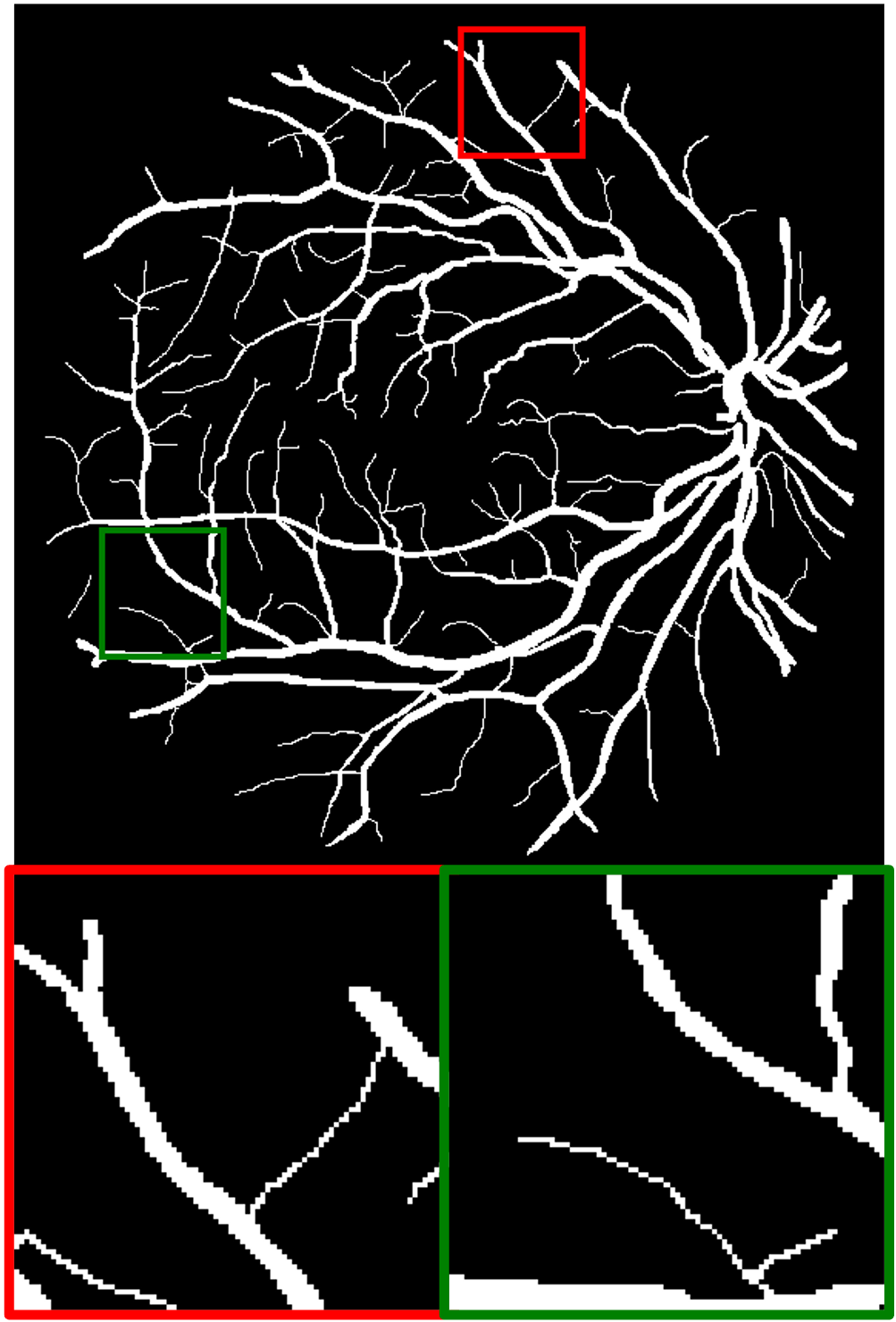}
        \par\vspace{0.5em}
        \caption{Ground Truth}
    \end{subfigure}
    \hfill
    \begin{subfigure}[t]{0.23\textwidth}
        \centering
        \includegraphics[width=\linewidth]{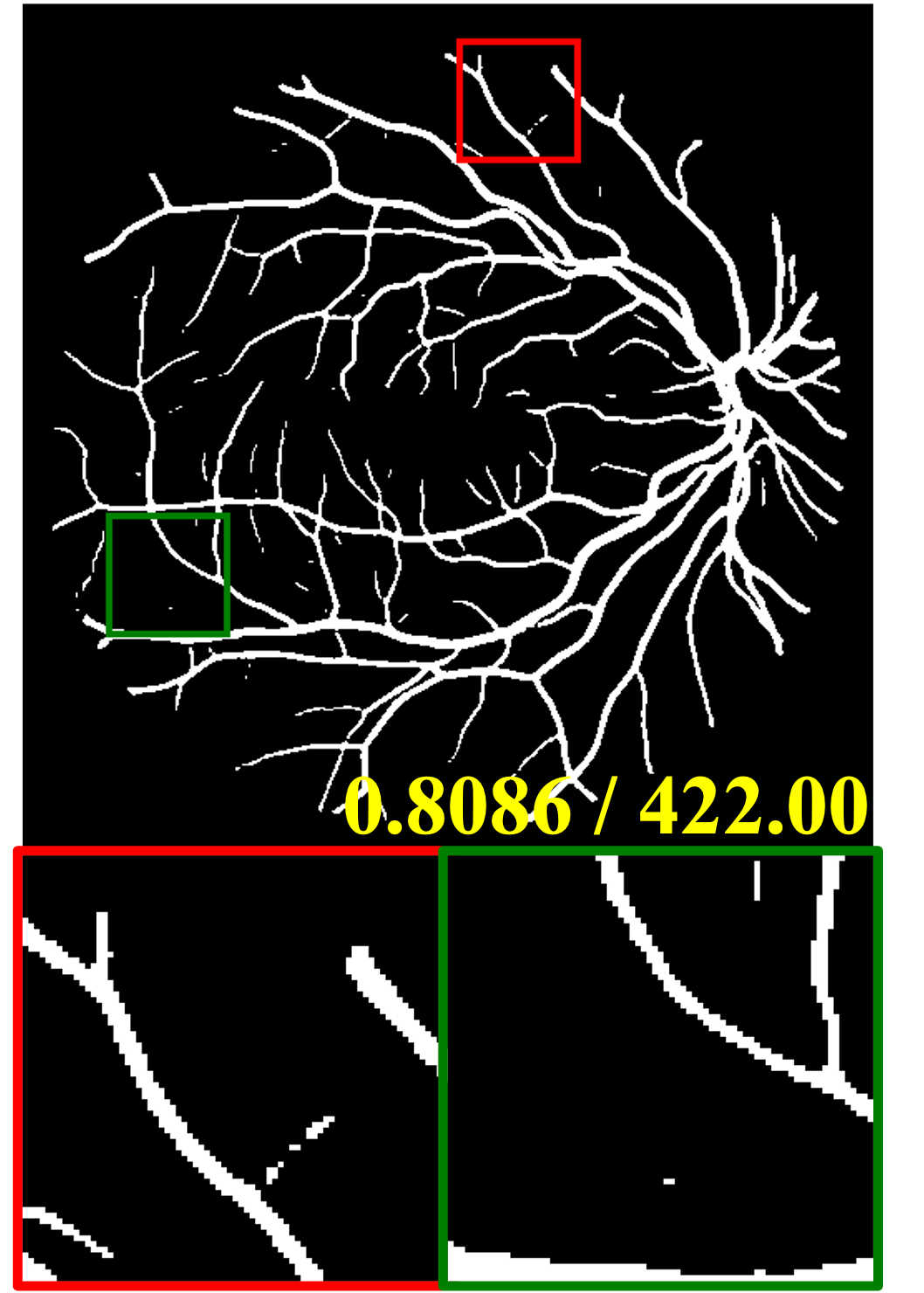}
        \par\vspace{0.5em}
        \caption{$Y$}
    \end{subfigure}
    \hfill
    \begin{subfigure}[t]{0.23\textwidth}
        \centering
        \includegraphics[width=\linewidth]{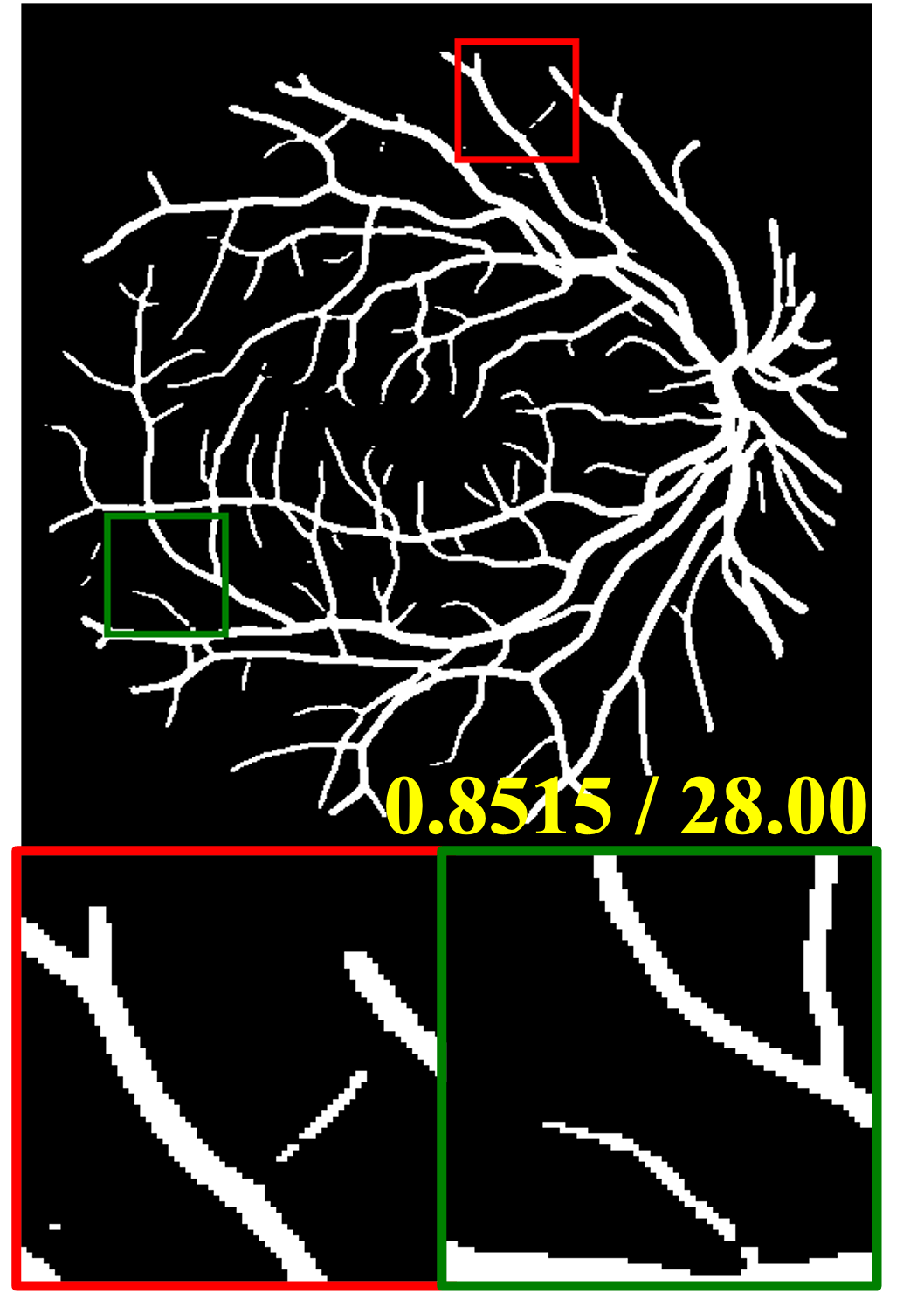}
        \par\vspace{0.5em}
        \caption{$\mathcal{D}$}
    \end{subfigure}
    \hfill
    \begin{subfigure}[t]{0.23\textwidth}
        \centering
        \includegraphics[width=\linewidth]{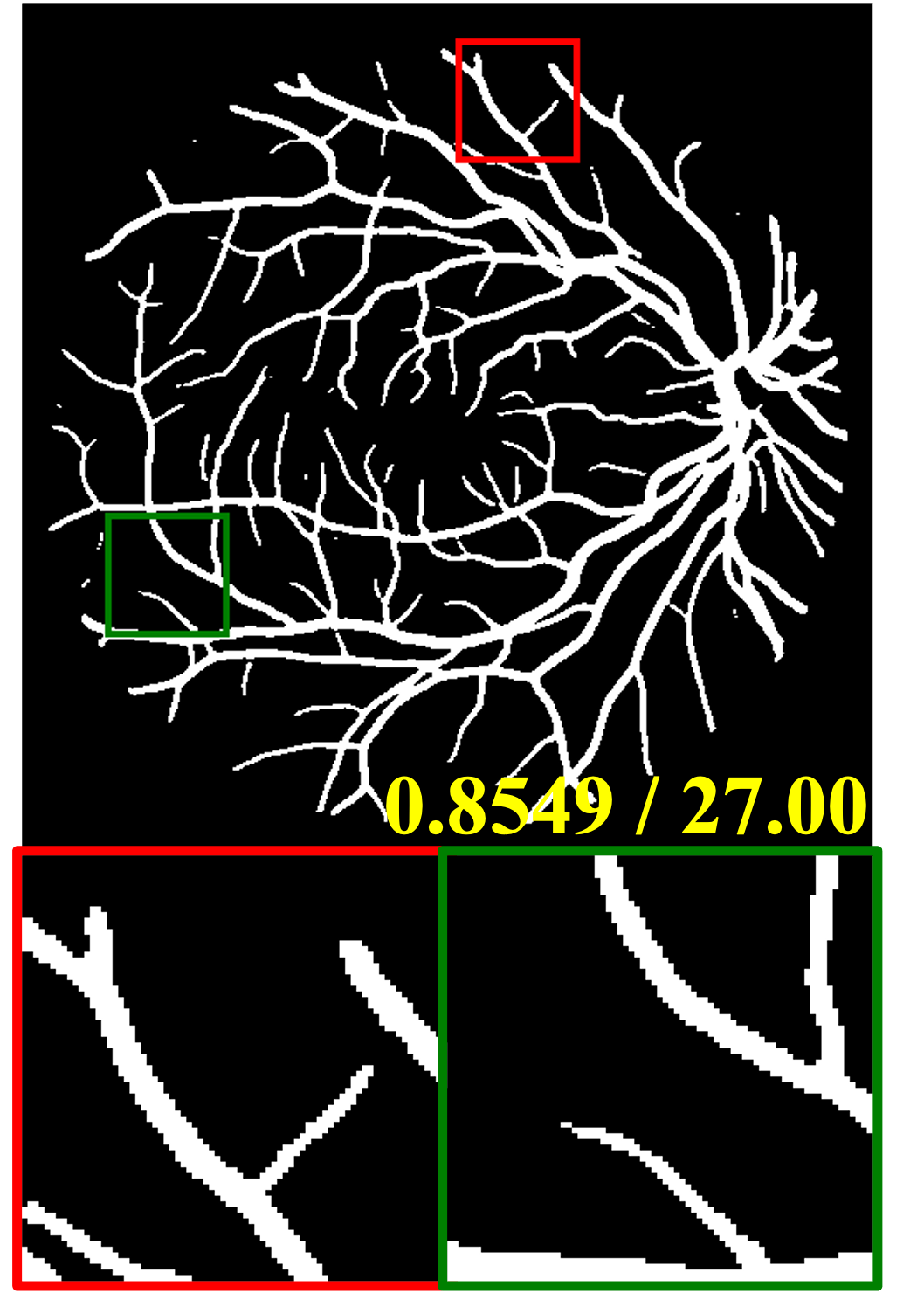}
        \par\vspace{0.5em}
        \caption{$\bar{\mathcal{D}}$\\ (Ours)}
    \end{subfigure}
    \caption{Qualitative comparison of the segmentation-label ablation on DRIVE\@.
        (a) Ground-truth mask.
        (b) to (d) predictions obtained using the binary ground-truth mask $Y$,
        the standard distance map $\mathcal{D}$, and the proposed inverted distance map $\bar{\mathcal{D}}$, respectively,
        as the source segmentation labels for affinity supervision.
        Red and green boxes mark the two regions enlarged underneath each panel.
        Yellow values give the Dice score and $\Delta\beta_0$.}
    \label{fig:sup_ablation_target}
\end{figure*}

\begin{figure*}[!t]
    \centering
    \captionsetup[subfigure]{skip=1pt,belowskip=-4pt}
    \begin{subfigure}[t]{0.18\textwidth}
        \centering
        \includegraphics[width=0.967\linewidth]{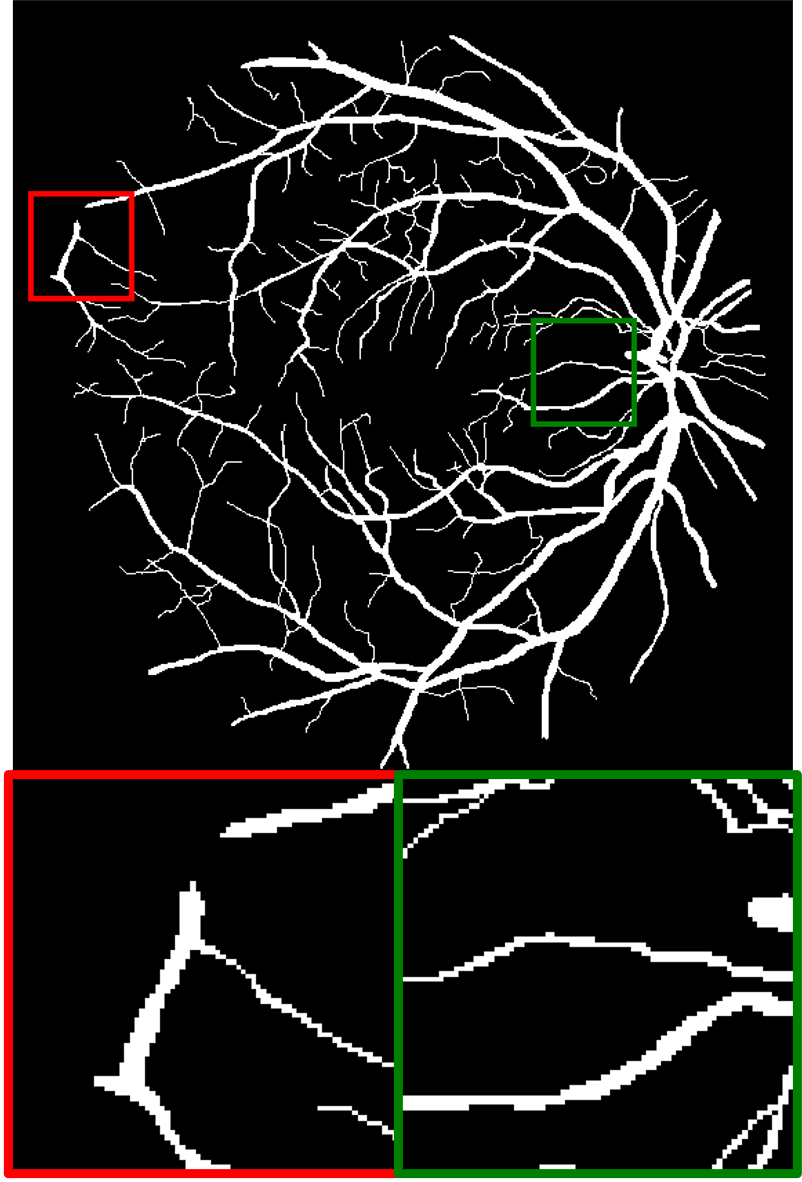}
        \par\vspace{0.5em}
        \caption{Ground Truth}
    \end{subfigure}
    \hfill
    \begin{subfigure}[t]{0.18\textwidth}
        \centering
        \includegraphics[width=\linewidth]{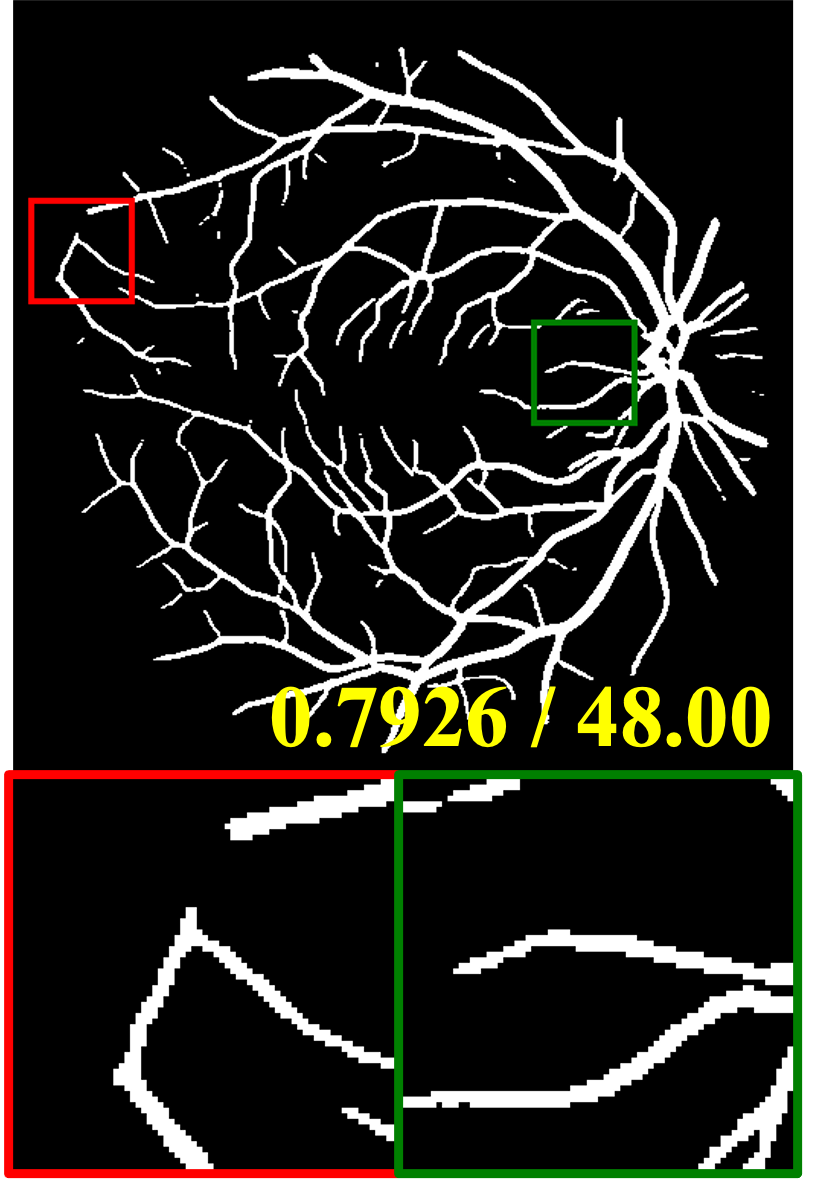}
        \par\vspace{0.5em}
        \caption{$\mathcal{S}=\{1\}$}
    \end{subfigure}
    \hfill
    \begin{subfigure}[t]{0.18\textwidth}
        \centering
        \includegraphics[width=\linewidth]{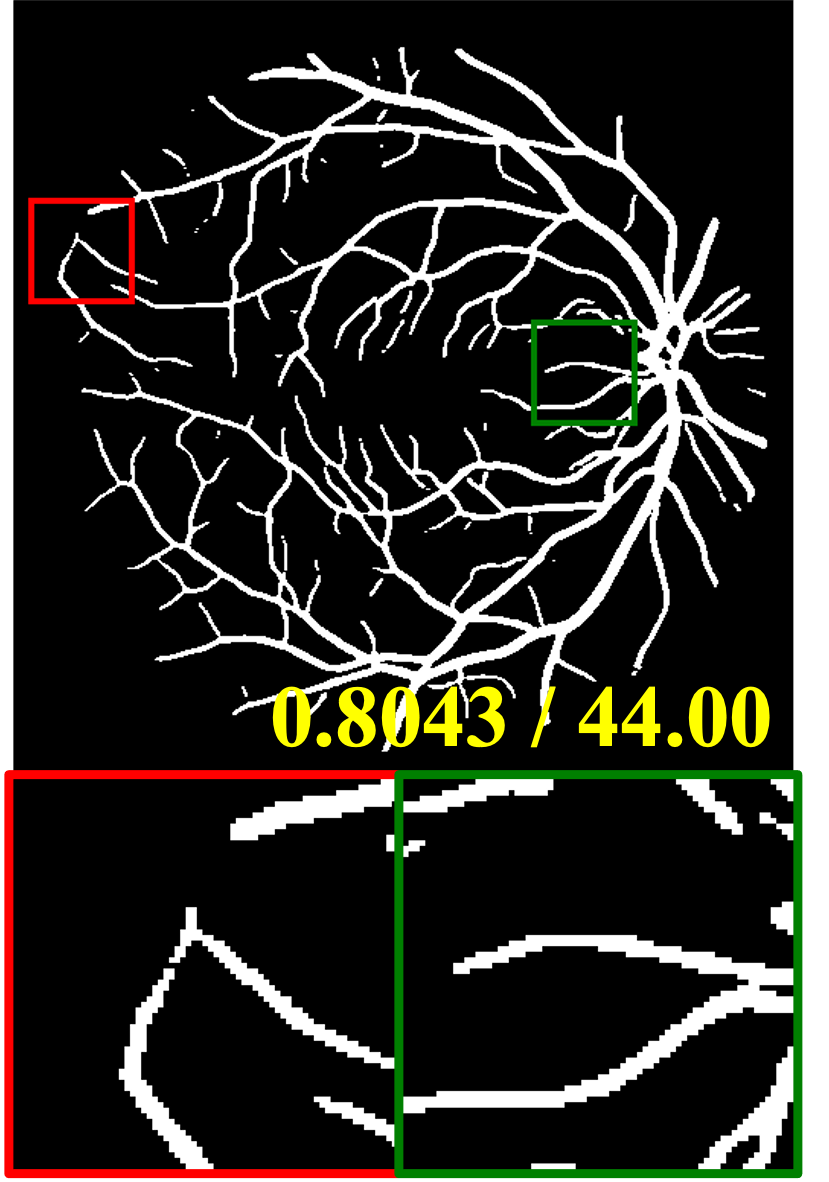}
        \par\vspace{0.5em}
        \caption{$\mathcal{S}=\{1,2\}$\\ (Ours)}
    \end{subfigure}
    \hfill
    \begin{subfigure}[t]{0.18\textwidth}
        \centering
        \includegraphics[width=0.995\linewidth]{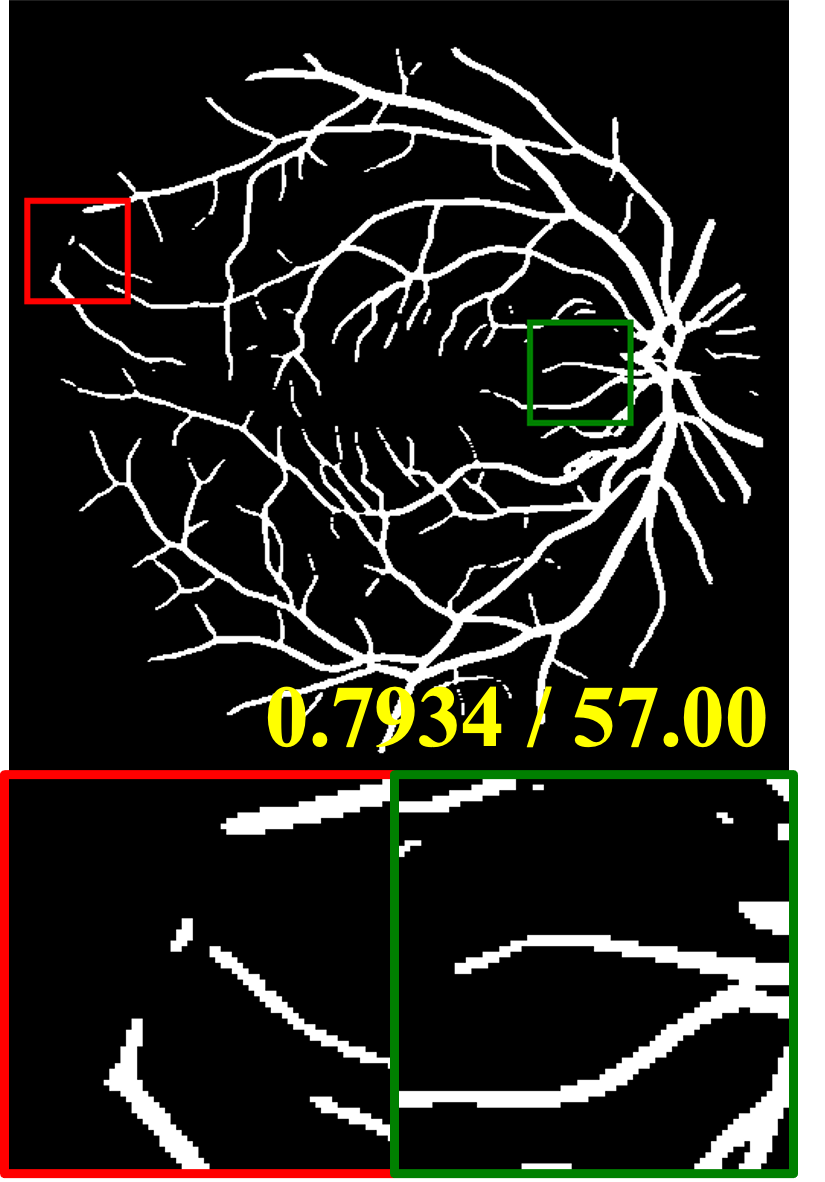}
        \par\vspace{0.5em}
        \caption{$\mathcal{S}=\{1,2^{\prime}\}$}
    \end{subfigure}
    \hfill
    \begin{subfigure}[t]{0.18\textwidth}
        \centering
        \includegraphics[width=0.995\linewidth]{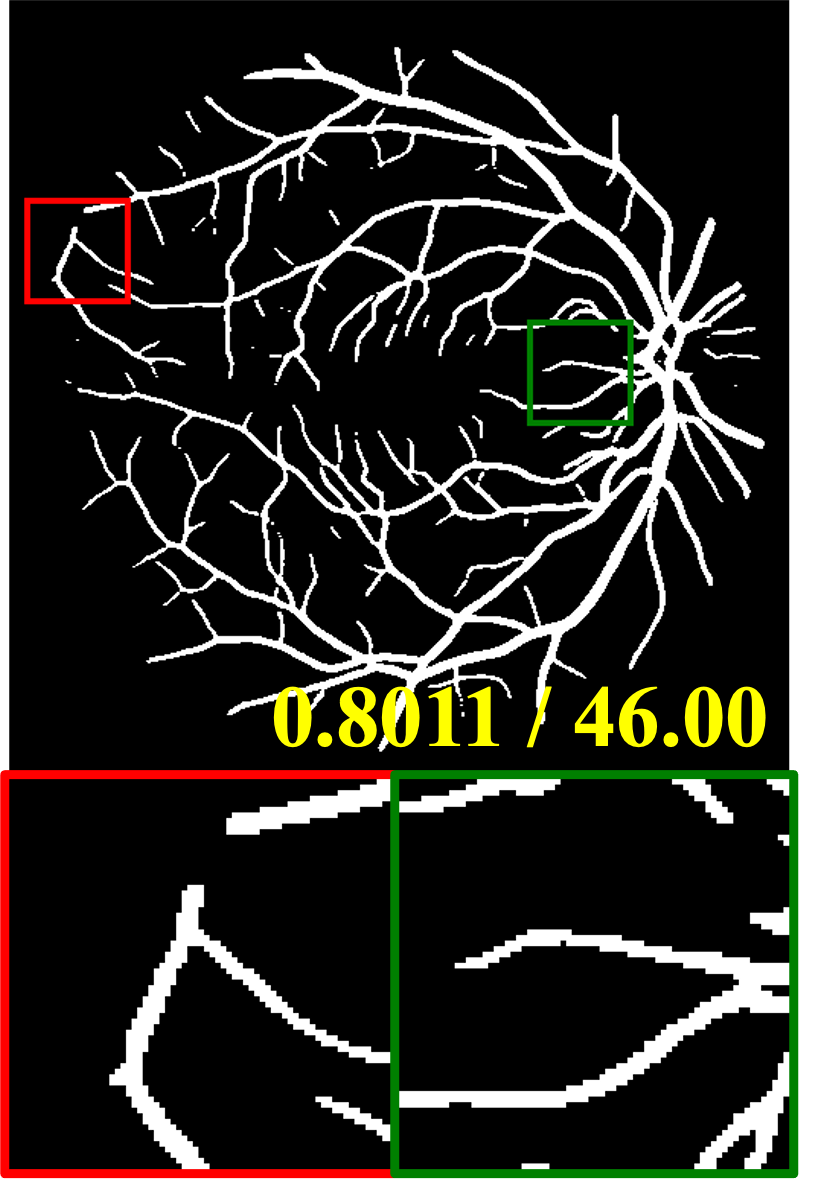}
        \par\vspace{0.5em}
        \caption{$\mathcal{S}=\{1,2,2^{\prime}\}$}
    \end{subfigure}
    \caption{Qualitative comparison of the branch-set ablation on DRIVE, varying the direction set at offset~2.
    (a) Ground-truth mask.
    (b) to (e) predictions with $\mathcal{S}=\{1\}$, $\mathcal{S}=\{1,2\}$,
    $\mathcal{S}=\{1,2^{\prime}\}$ and $\mathcal{S}=\{1,2,2^{\prime}\}$, respectively.
    The red and green boxes mark the two regions enlarged underneath each panel,
    and the yellow values give the Dice score and $\Delta\beta_0$.
    The best configuration $\mathcal{S}=\{1,2\}$ is labeled (Ours) in the figure.}
    \label{fig:sup_ablation_direction}
\end{figure*}

\begin{figure*}[!t]
    \centering
    \captionsetup[subfigure]{skip=1pt,belowskip=-4pt}
    \begin{subfigure}[t]{0.18\textwidth}
        \centering
        \includegraphics[width=0.962\linewidth]{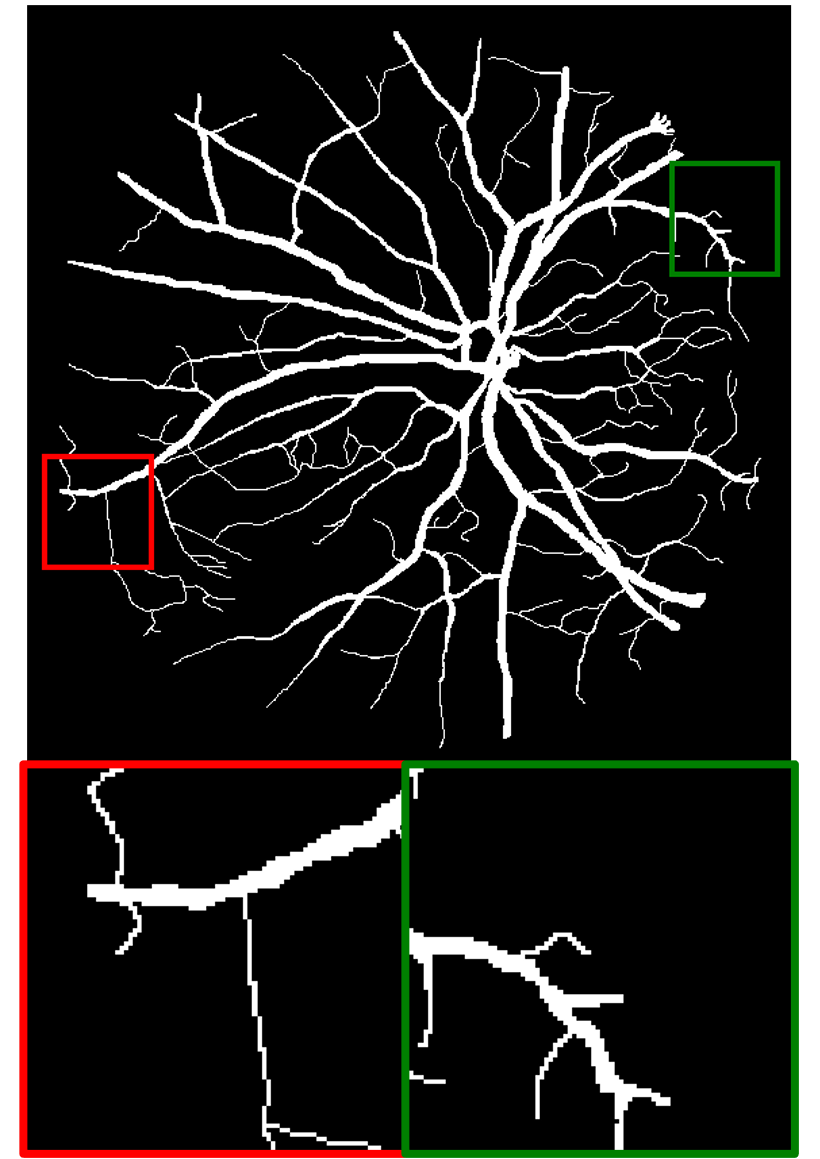}
        \par\vspace{0.5em}
        \caption{Ground Truth}
    \end{subfigure}
    \hfill
    \begin{subfigure}[t]{0.18\textwidth}
        \centering
        \includegraphics[width=0.989\linewidth]{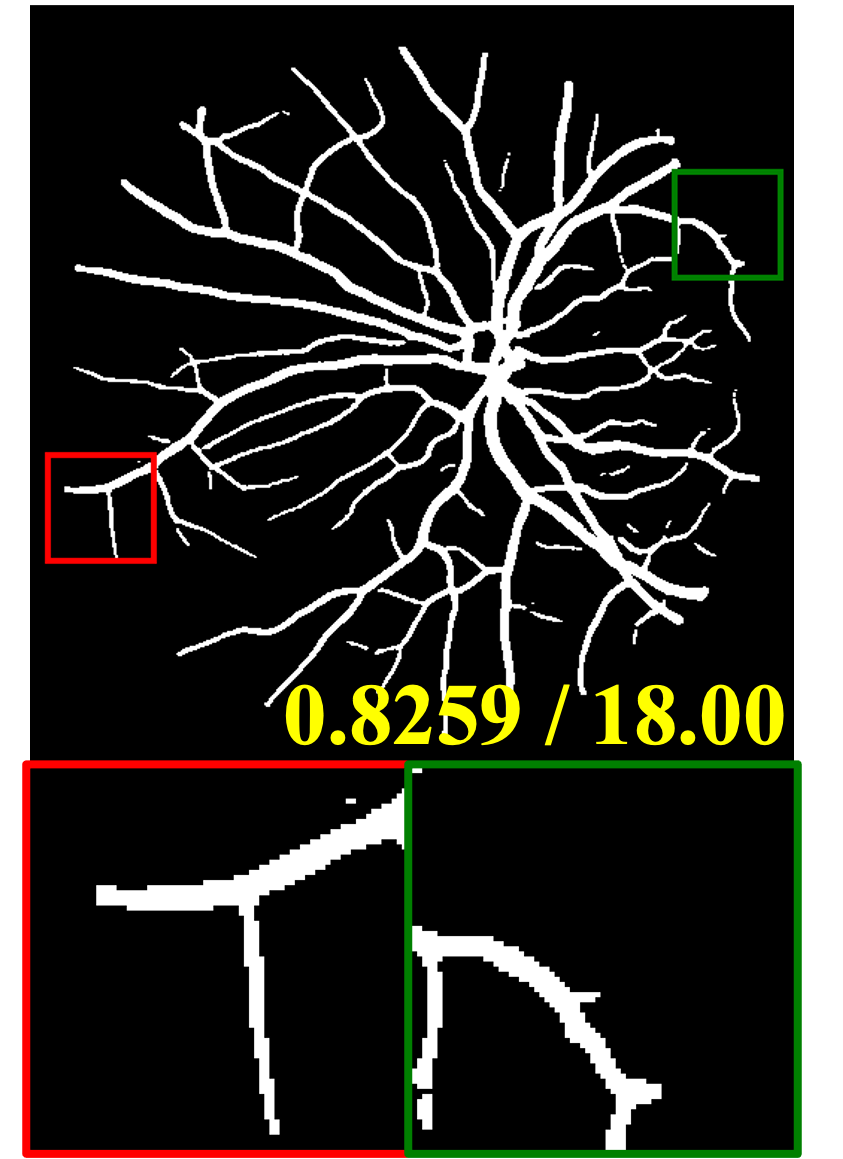}
        \par\vspace{0.5em}
        \caption{$\mathcal{S}=\{1\}$}
    \end{subfigure}
    \hfill
    \begin{subfigure}[t]{0.18\textwidth}
        \centering
        \includegraphics[width=0.979\linewidth]{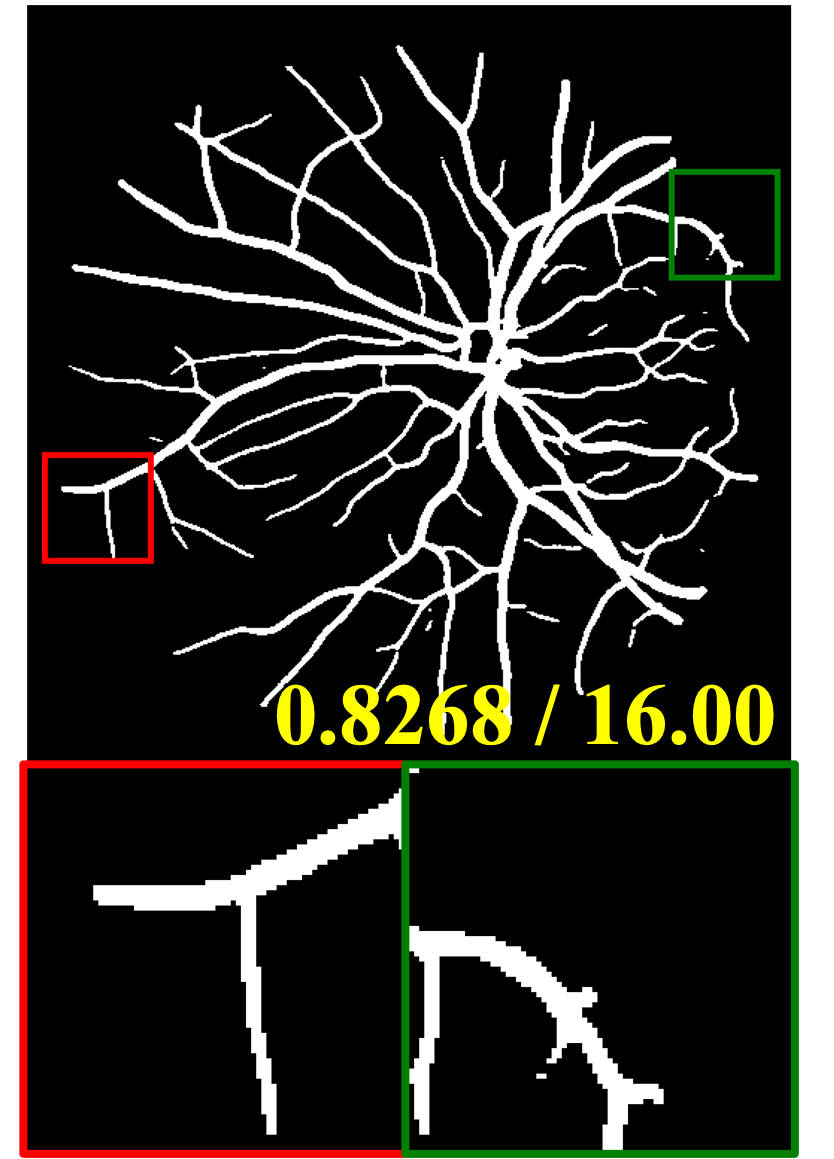}
        \par\vspace{0.5em}
        \caption{$\mathcal{S}=\{1,2\}$\\ (Ours)}
    \end{subfigure}
    \hfill
    \begin{subfigure}[t]{0.18\textwidth}
        \centering
        \includegraphics[width=\linewidth]{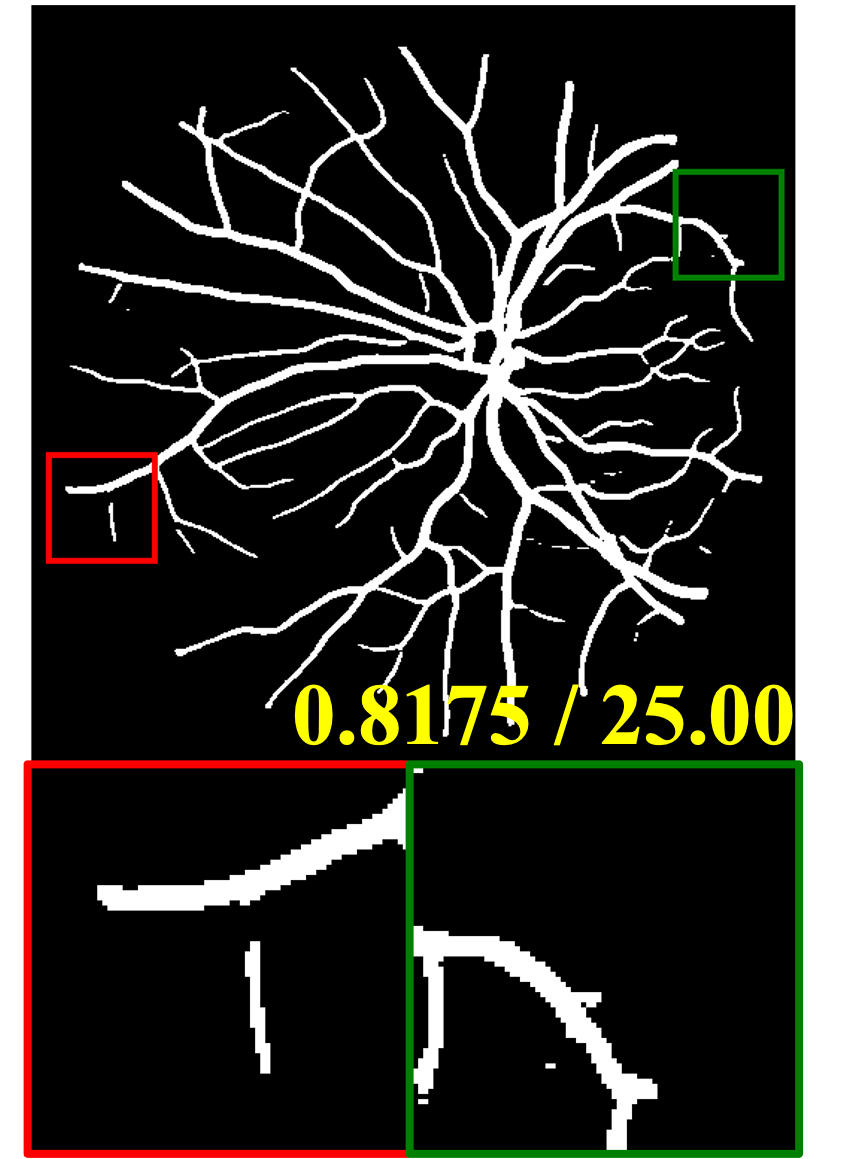}
        \par\vspace{0.5em}
        \caption{$\mathcal{S}=\{1,2,3\}$}
    \end{subfigure}
    \hfill
    \begin{subfigure}[t]{0.18\textwidth}
        \centering
        \includegraphics[width=0.994\linewidth]{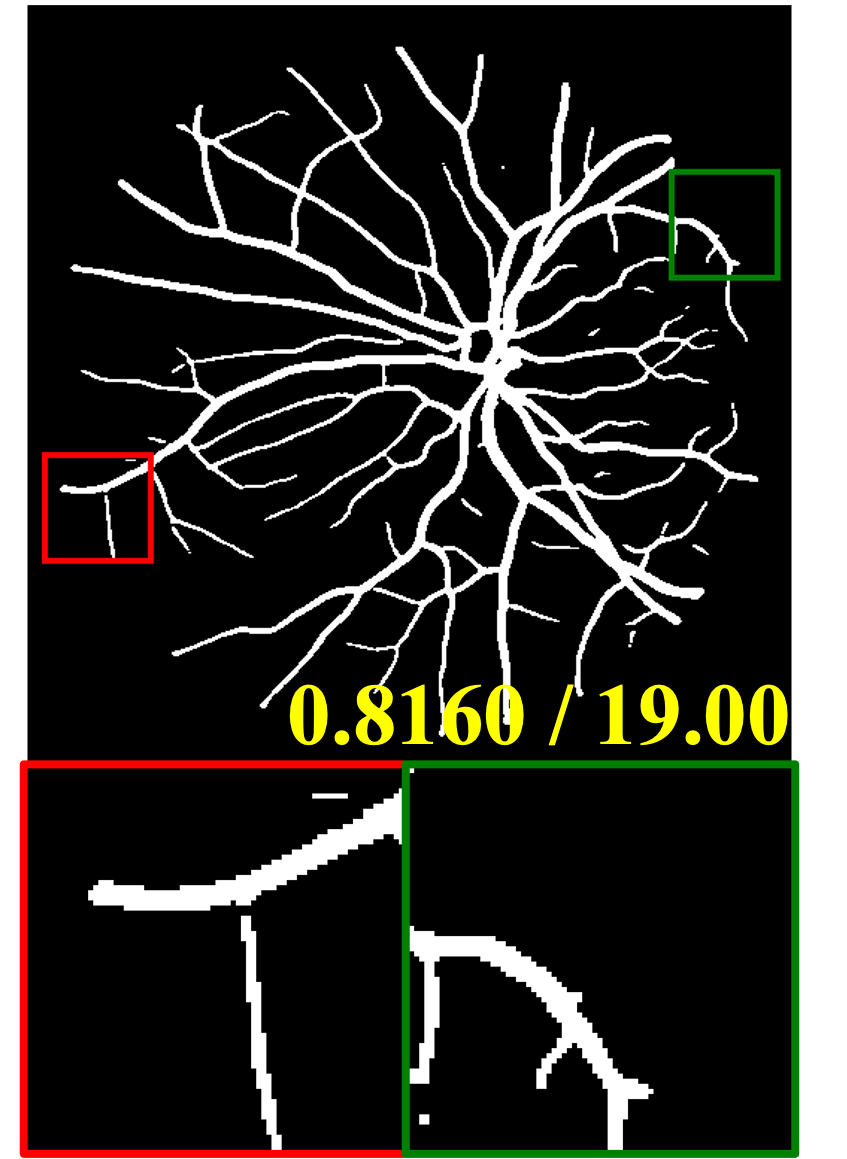}
        \par\vspace{0.5em}
        \caption{$\mathcal{S}=\{1,2,3,4\}$}
    \end{subfigure}
    \caption{Qualitative comparison of the branch-set ablation on DRIVE, varying the number of sampling offsets.
        (a) Ground-truth mask.
        (b) to (e) predictions with $\mathcal{S}=\{1\}$, $\mathcal{S}=\{1,2\}$,
        $\mathcal{S}=\{1,2,3\}$ and $\mathcal{S}=\{1,2,3,4\}$, respectively.
        The red and green boxes mark the two regions enlarged underneath each panel,
        and the yellow values give the Dice score and $\Delta\beta_0$.
        The best configuration $\mathcal{S}=\{1,2\}$ is labeled (Ours) in the figure.}
    \label{fig:sup_ablation_scale}
\end{figure*}

\begin{figure}[!t]
    \centering
    \includegraphics[width=\columnwidth]{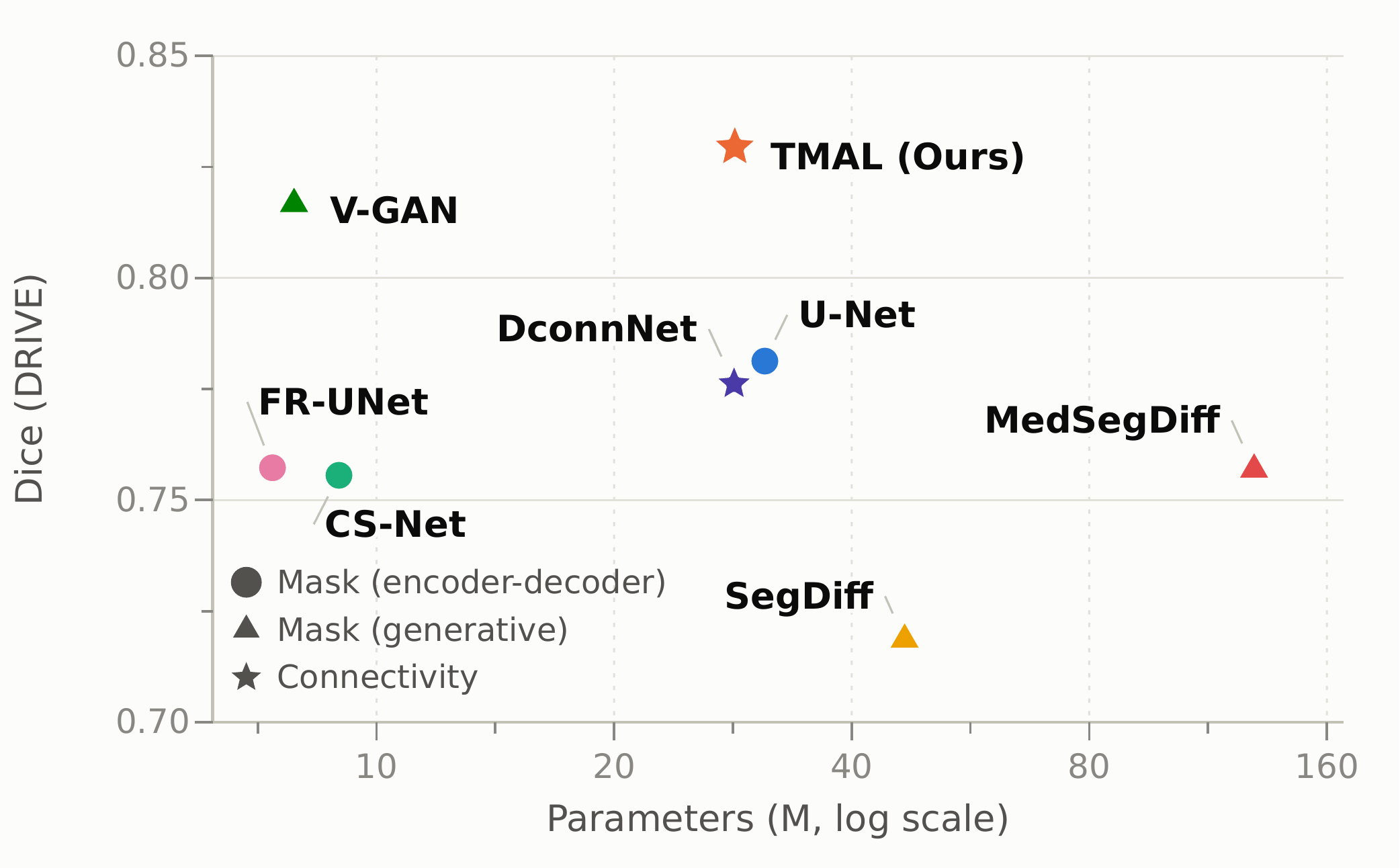}
    \caption{Parameter count against Dice score on DRIVE for TMAL and representative baselines.
        Marker shape follows the grouping of Table~1 of the main paper, with the mask-based methods split into the encoder-decoder and the generative ones.
        Topology-based methods are losses added to an existing backbone, so their parameter counts equal those of that backbone and they are not plotted separately.
        All counts are total parameters under one shared setup.}
    \label{fig:sup_param}
\end{figure}

This section gives the formal definitions of the evaluation metrics reported in the
main paper. Let $Y\in\{0,1\}^{H\times W}$ denote the ground-truth mask and
$\hat{Y}\in\{0,1\}^{H\times W}$ the predicted mask, and let $\Omega$ denote the
image domain.
The pixel-level true positives, false positives, and false negatives are
\begin{equation}\label{eq:sup_confusion}
    \begin{aligned}
        \mathrm{TP} & = \lvert Y\cap\hat{Y}\rvert,                   \\
        \mathrm{FP} & = \lvert (\Omega\setminus Y)\cap\hat{Y}\rvert, \\
        \mathrm{FN} & = \lvert Y\cap(\Omega\setminus\hat{Y})\rvert.
    \end{aligned}
\end{equation}

\paragraph{Dice and IoU}
\begin{equation}\label{eq:sup_metric_dice}
    \mathrm{Dice}
    =
    \frac{2\,\mathrm{TP}}{2\,\mathrm{TP}+\mathrm{FP}+\mathrm{FN}},
\end{equation}
\begin{equation}\label{eq:sup_iou}
    \mathrm{IoU}
    =
    \frac{\mathrm{TP}}{\mathrm{TP}+\mathrm{FP}+\mathrm{FN}}.
\end{equation}
Dice measures foreground overlap, whereas IoU measures the proportion of the
ground-truth--prediction union that is correctly predicted as foreground.

\paragraph{clDice}
clDice~\cite{S-shit2021cldice} measures topology-aware overlap between the prediction and ground truth through their morphological skeletons.
Let $\operatorname{Skel}(\cdot)$ denote the morphological skeletonization operator, implemented using \texttt{skimage.morphology.skeletonize}.
The topology precision and sensitivity are
\begin{equation}\label{eq:sup_tprec}
    \begin{aligned}
        T_{\mathrm{prec}}(Y,\hat{Y})
         & =
        \frac{\lvert Y\cap \operatorname{Skel}(\hat{Y})\rvert}
        {\lvert\operatorname{Skel}(\hat{Y})\rvert}, \\
        T_{\mathrm{sens}}(Y,\hat{Y})
         & =
        \frac{\lvert \hat{Y}\cap \operatorname{Skel}(Y)\rvert}
        {\lvert\operatorname{Skel}(Y)\rvert}.
    \end{aligned}
\end{equation}
The clDice score is their harmonic mean:
\begin{equation}\label{eq:sup_cldice}
    \mathrm{clDice}(Y,\hat{Y})
    =
    \frac{2\,T_{\mathrm{prec}}\,T_{\mathrm{sens}}}{T_{\mathrm{prec}}+T_{\mathrm{sens}}}.
\end{equation}

\paragraph{Betti-number Error}
Betti numbers characterize topology in a binary mask~\cite{S-stucki2023topologically}.
$\beta_0$ is the number of separate foreground components.
A break in a vessel or crack can split one structure into several components and thus increase $\beta_0$.
$\beta_1$ is the number of enclosed holes. An unwanted closed loop can create an extra hole and increase $\beta_1$.
\begin{equation}\label{eq:sup_betti_error}
    \Delta\beta_k
    =
    \left|\beta_k\!\left(\hat{Y}\right)
    -\beta_k\!\left(Y\right)\right|.
\end{equation}
A zero error means that the prediction and ground truth have the same number of components or holes.
However, this metric compares only the counts, not their locations, so a missing structure and an extra structure can cancel each other~\cite{S-stucki2023topologically}.
We therefore interpret it together with clDice and HD95.

\Needspace{14\baselineskip}
\paragraph{Hausdorff Distance}
The Hausdorff distance measures separation between the predicted and ground-truth boundaries.
Let $\partial Y$ and $\partial\hat{Y}$ denote their boundary point sets.
For boundary point sets $A$ and $B$, we define the directed and symmetric Hausdorff distances as
\begin{align}
    h(A,B)
     & =
    \max_{a\in A}\min_{b\in B}\|a-b\|,
    \label{eq:sup_directed_hausdorff} \\
    \mathrm{HD}(Y,\hat{Y})
     & =
    \max\bigl\{h(\partial\hat{Y},\partial Y),\,
    h(\partial Y,\partial\hat{Y})\bigr\}.
    \label{eq:sup_hausdorff}
\end{align}
We report HD95, replacing the outer maximum with the 95th percentile of point-to-boundary distances to reduce sensitivity to outliers.
Lower values indicate better boundary agreement.

\section{Additional Analysis and Results}
\label{sec:sup_qual}

The figures in this section illustrate the ablation table of the main paper.
Figure~\ref{fig:sup_ablation_target} corresponds to its segmentation-label block, whereas
Figure~\ref{fig:sup_ablation_direction} and Figure~\ref{fig:sup_ablation_scale} both belong to its branch-set block,
the former varying the direction set at offset~2 and the latter the number of sampling offsets.
The objective-term block is quantitative only and has no counterpart here.
The sampling patterns that these branch sets use are drawn in Figure~4 of the main paper.

\Needspace{9\baselineskip}
Figure~\ref{fig:sup_param} places TMAL against the baselines in the Dice-versus-params plane on DRIVE\@.
TMAL has $28.44$\,M parameters against the $28.38$\,M of DconnNet~\cite{S-yang2023directional}, the closest connectivity-based baseline,
so TMAL uses about $0.2\%$ more parameters.
Thus, the reported gains are not accompanied by a substantial increase in parameter count.
For context, U-Net~\cite{S-ronneberger2015u} has a comparable parameter count of $31.04$\,M, whereas MedSegDiff~\cite{S-wu2024medsegdiff}, the largest model in the comparison, has $129.40$\,M.

{
\small
\interlinepenalty=10000

}

\end{document}